# A Survey of the Recent Architectures of Deep Convolutional Neural Networks

Asifullah Khan[1, 2*], Anabia Sohail[1, 2], Umme Zahoora[1], and Aqsa Saeed Qureshi[1]

[1] Pattern Recognition Lab, DCIS, PIEAS, Nilore, Islamabad 45650, Pakistan
[2] Deep Learning Lab, Center for Mathematical Sciences, PIEAS, Nilore, Islamabad 45650, Pakistan
asif@pieas.edu.pk

## Abstract

Deep Convolutional Neural Network (CNN) is a special type of Neural Networks, which has shown exemplary performance on several competitions related to Computer Vision and Image Processing. Some of the exciting application areas of CNN include Image Classification and Segmentation, Object Detection, Video Processing, Natural Language Processing, and Speech Recognition. The powerful learning ability of deep CNN is primarily due to the use of multiple feature extraction stages that can automatically learn representations from the data. The availability of a large amount of data and improvement in the hardware technology has accelerated the research in CNNs, and recently interesting deep CNN architectures have been reported. Several inspiring ideas to bring advancements in CNNs have been explored, such as the use of different activation and loss functions, parameter optimization, regularization, and architectural innovations. However, the significant improvement in the representational capacity of the deep CNN is achieved through architectural innovations. Notably, the ideas of exploiting spatial and channel information, depth and width of architecture, and multi-path information processing have gained substantial attention. Similarly, the idea of using a block of layers as a structural unit is also gaining popularity. This survey thus focuses on the intrinsic taxonomy present in the recently reported deep CNN architectures and, consequently, classifies the recent innovations in CNN architectures into seven different categories. These seven categories are based on spatial exploitation, depth, multi-path, width, feature-map exploitation, channel boosting, and attention. Additionally, the elementary understanding of CNN components, current challenges, and applications of CNN are also provided.

# 1   Introduction

Machine Learning (ML) algorithms are known to learn the underlying relationship in data and thus make decisions without requiring explicit instructions. In literature, various exciting works have been reported to understand and/or emulate the human sensory responses such as speech and vision (Hubel and Wiesel 1962, 1968; Ojala et al. 1996; Chapelle 1998; Lowe 1999; Dalal and Triggs 2004; Bay et al. 2008; Heikkilä et al. 2009). In 1989, a new class of Neural Networks (NN), called Convolutional Neural Network (CNN) (LeCun et al. 1989) was reported, which has shown enormous potential in Machine Vision (MV) related tasks.

CNNs are one of the best learning algorithms for understanding image content and have shown exemplary performance in image segmentation, classification, detection, and retrieval related tasks (Ciresan et al. 2012; Liu et al. 2019). The success of CNNs has captured attention beyond academia. In industry, companies such as Google, Microsoft, AT&T, NEC, and Facebook have developed active research groups for exploring new architectures of CNN (Deng et al. 2013). At present, most of the frontrunners of image processing and computer vision (CV) competitions are employing deep CNN based models.

The attractive feature of CNN is its ability to exploit spatial or temporal correlation in data. The topology of CNN is divided into multiple learning stages composed of a combination of the convolutional layers, non-linear processing units, and subsampling layers (Jarrett et al. 2009). CNN is a feedforward multilayered hierarchical network, where each layer, using a bank of convolutional kernels, performs multiple transformations (LeCun et al. 2010). Convolution operation helps in the extraction of useful features from locally correlated data points. The output of the convolutional kernels is then assigned to the non-linear processing unit (activation function), which not only helps in learning abstractions but also embeds non-linearity in the feature space. This non-linearity generates different patterns of activations for different responses and thus facilitates in learning of semantic differences in images. The output of the non-linear activation function is usually followed by subsampling, which helps in summarizing the results and also makes the input invariant to geometrical distortions (Scherer et al. 2010; LeCun et al. 2010). CNN, with the automatic feature extraction ability, reduces the need for a separate feature extractor (Najafabadi et al. 2015). Thus, CNN without exhaustive processing can learn good internal representation from raw pixels. Notable attributes of CNN are hierarchical learning, automatic feature extraction, multi-tasking, and weight sharing (Guo et al. 2016; Liu et al. 2017; Abbas et al. 2019).

CNN first came to limelight through the work of LeCuN in 1989 for processing of grid-like topological data (images and time series data) (LeCun et al. 1989; Ian Goodfellow et al. 2017). The architectural design of CNN was inspired by Hubel and Wiesel's work and thus mostly follows the basic structure of primate's visual cortex (Hubel and Wiesel 1962, 1968). Different stages of the learning process in CNN show quite a resemblance to the primate's ventral pathway of the visual cortex (V1-V2-V3-V4-IT/VTC) (Laskar et al. 2018). The visual cortex of primates first receives input from the retinotopic area. Whereby, the lateral geniculate nucleus performs multi-scale highpass filtering and contrast normalization. After this, detection is performed by different regions of the visual cortex categorized as V1, V2, V3, and V4. In fact, V1 and V2 regions of the visual cortex are similar to convolutional and subsampling layers. In contrast, the inferior temporal region resembles the higher layers of CNN, which makes an inference about the image (Grill-Spector et al. 2018).

During training, CNN learns through backpropagation algorithm, by regulating the change in weights according to the target. Optimization of an objective function using a backpropagation algorithm is similar to the response based learning of the human brain. The multilayered, hierarchical structure of deep CNN, gives it the ability to extract low, mid, and high-level features. High-level features (more abstract features) are a combination of lower and mid-level features. The hierarchical feature extraction ability of CNN emulates the deep and layered learning process of the Neocortex in the human brain, which dynamically learns features from the raw data (Bengio 2009). The popularity of CNN is primarily due to its hierarchical feature extraction ability.

Deep architectures often have an advantage over shallow architectures when dealing with complex learning problems. The stacking of multiple linear and non-linear processing units in a layer-wise fashion provides the ability to learn complex representations at different levels of abstraction. Consequently, in recognition tasks consisting of hundreds of image categories, deep CNNs have shown substantial performance improvement over conventional vision-based models (Ojala et al. 2002; Dalal and Triggs 2004; Lowe 2004). The observation that the deep architectures can improve the representational capacity of a CNN heightened the use of CNN in image classification and segmentation tasks (Krizhevsky et al. 2012). The availability of big data and advancements in hardware are also the main reasons for the recent success of deep CNNs.

Empirical studies showed that if given enough training data, deep CNNs can learn the invariant representations and may achieve human-level performance. In addition to its use as a supervised learning mechanism, the potential of deep CNNs can also be exploited to extract useful representations from a large scale of unlabeled data. Recently, it is shown that different levels of features, including both low and high-level, can be transferred to a generic recognition task by exploiting the concept of Transfer Learning (TL) (Qiang Yang et al. 2008; Qureshi et al. 2017; Qureshi and Khan 2018).

From the late 1990s up to 2000, various improvements in CNN learning methodology and architecture were performed to make CNN scalable to large, heterogeneous, complex, and multiclass problems. Innovations in CNNs include different aspects such as modification of processing units, parameter and hyper-parameter optimization strategies, design patterns and connectivity of layers, etc. CNN based applications became prevalent after the exemplary performance of AlexNet on the ImageNet dataset in 2012 (Krizhevsky et al. 2012). Significant innovations in CNN have been proposed since then and are largely attributed to the restructuring of processing units and designing of new blocks. Zeiler and Fergus (Zeiler and Fergus 2013) gave the concept of layer-wise visualization of CNN to improve the understanding of feature extraction stages, which shifted the trend towards extraction of features at low spatial resolution in deep architecture as performed in VGG (Simonyan and Zisserman 2015). Nowadays, most of the new architectures are built upon the principle of simple and homogenous topology, as introduced in VGG. Google deep learning group introduced an innovative idea of a split, transform and merge, with the corresponding block known as inception block. The inception block for the very first time gave the concept of branching within a layer, which allows abstraction of features at different spatial scales (Szegedy et al. 2015). In 2015, the concept of skip connections introduced by ResNet (He et al. 2015a) for the training of deep CNNs gained popularity. Afterward, this concept was used by most of the succeeding networks, such as Inception-ResNet, Wide ResNet, ResNeXt, etc., (Szegedy et al. 2016a; Zagoruyko and Komodakis 2016; Xie et al. 2017).

Different architectural designs such as Wide ResNet, ResNeXt, Pyramidal Net, Xception, PolyNet, and many others explore the effect of multilevel transformations on CNNs learning capacity by introducing cardinality or increasing the width (Zagoruyko and Komodakis 2016; Han et al. 2017; Xie et al. 2017; Zhang et al. 2017). Therefore, the focus of research shifted from parameter optimization and connections readjustment towards the improved architectural design of the network. This shift resulted in many new architectural ideas such as channel boosting, spatial and feature-map wise exploitation and attention-based information processing etc., (Wang et al. 2017a; Khan et al. 2018a; Woo et al. 2018).

In the past few years, different interesting surveys are conducted on deep CNNs that elaborate on the essential components of CNN and their alternatives. The survey reported in (Gu et al. 2018) reviewed the famous architectures from 2012-2015 along with their basic components. Similarly, there are prominent surveys that discuss different algorithms and applications of CNN (LeCun et al. 2010; Najafabadi et al. 2015; Guo et al. 2016; Srinivas et al. 2016; Liu et al. 2017). Likewise, the survey presented in (Zhang et al. 2019) discusses the taxonomy of CNNs based on acceleration techniques. On the other hand, in this survey, we discuss the intrinsic taxonomy present in the recent and prominent CNN architectures reported from 2012-2020. The various CNN architectures discussed in this survey are broadly classified into seven main categories, namely; spatial exploitation, depth, multi-path, width, feature-map exploitation, channel boosting, and attention-based CNNs.

This survey also gives an insight into the basic structure of CNN as well as its historical perspective, presenting different eras of CNN that trace back from its origin to its latest developments and achievements. This survey will help the readers to develop the theoretical insight into the design principles of CNN and thus may further accelerate the architectural innovations in CNN.

The rest of the paper is organized in the following order (shown in Fig. 1): Section 1 develops the systematic understanding of CNN, discusses its resemblance with primate's visual cortex, as well as its contribution to MV. In this regard, Section 2 provides an overview of essential CNN components, and Section 3 discusses the architectural evolution of deep CNNs. Whereas, Section 4 discusses the recent innovations in CNN architectures and categorizes CNNs into seven broad classes. Section 5 and 6 shed light on applications of CNNs and current challenges, whereas section 7 discusses future work. Finally, the last section concludes.

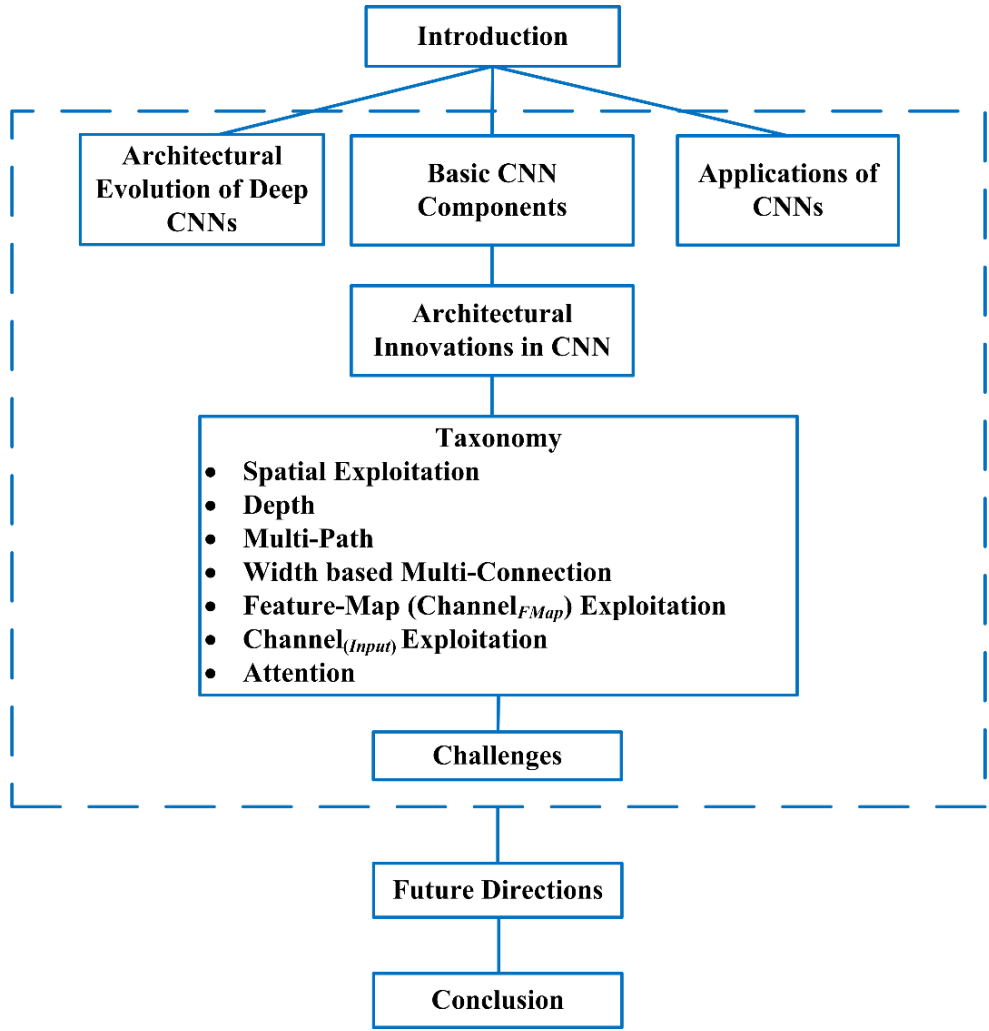

**Fig. 1** Organization of the survey paper showing different sections.

## 2 Basic CNN components

Nowadays, CNN is considered as one of the most widely used ML technique, especially in vision-related applications. CNN can learn representations from the grid-like data, and recently it has shown substantial performance improvement in various ML applications. A typical block diagram of an ML system is shown in Fig. 2. Since CNN possesses both good feature generation and discrimination ability, therefore in a typical ML system, CNN capabilities are exploited for feature generation and classification.

A typical CNN architecture generally comprises alternate layers of convolution and pooling followed by one or more fully connected layers at the end. In some cases, a fully connected layer is replaced with a global average pooling layer. In addition to different mapping functions, different regulatory units such as batch normalization and dropout are also incorporated to optimize CNN performance (Bouvrie 2006). The arrangement of CNN components plays a fundamental role in designing new architectures and thus achieving enhanced performance. This section briefly discusses the role of these components in a CNN architecture.

## 2.1    Convolutional layer

The convolutional layer is composed of a set of convolutional kernels where each neuron acts as a kernel. However, if the kernel is symmetric, the convolution operation becomes a correlation operation (Ian Goodfellow et al. 2017). Convolutional kernel works by dividing the image into small slices, commonly known as receptive fields. The division of an image into small blocks helps in extracting feature motifs. Kernel convolves with the images using a specific set of weights by multiplying its elements with the corresponding elements of the receptive field (Bouvrie 2006). Convolution operation can be expressed as follows:

$$f_l^k(p,q) = \sum_c \sum_{x,y} i_c(x,y).e_l^k(u,v) \tag{1}$$

where, $i_c(x,y)$ is an element of the input image tensor $I_C$, which is element wise multiplied by $e_l^k(u,v)$ index of the $k^{th}$ convolutional kernel $k_l$ of the $l^{th}$ layer. Whereas output feature-map of the $k^{th}$ convolutional operation can be expressed as $\mathbf{F}_l^k = \left[ f_l^k(1,1),...,f_l^k(p,q),...,f_l^k(P,Q) \right]$. The different mathematical symbols used are defined in Table 1.

Due to weight sharing ability of convolutional operation, different sets of features within an image can be extracted by sliding kernel with the same set of weights on the image and thus makes CNN parameter efficient as compared to the fully connected networks. Convolution operation may further be categorized into different types based on the type and size of filters, type of padding, and the direction of convolution (LeCun et al. 2015).

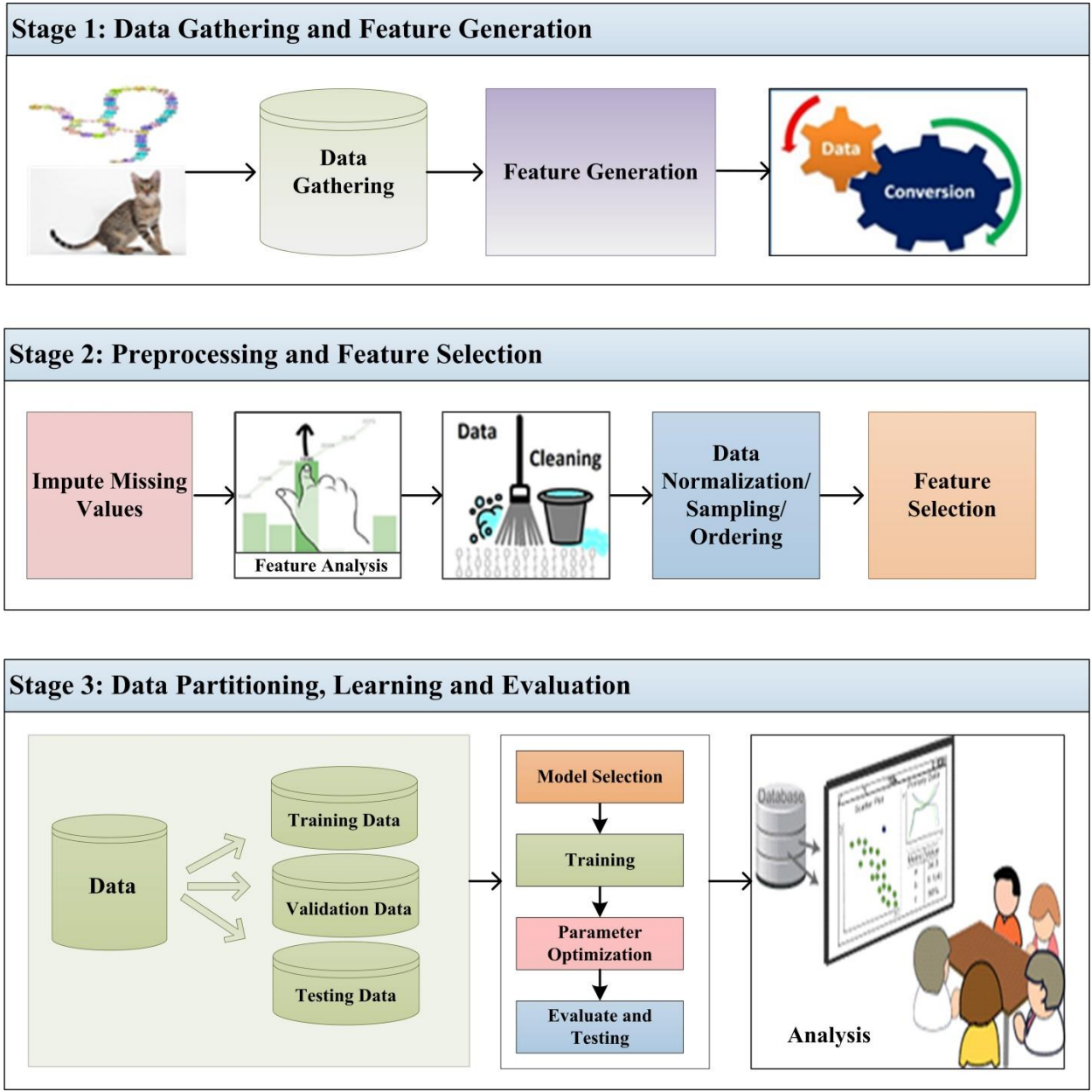

**Fig. 2** Basic layout of a typical ML system having several stages.

**Table 1** Definition of mathematical symbols

| Symbol | Description |
|---|---|
| $X$ | Total $x$ coordinates of an image |
| $x$ | $x^{th}$ coordinate under consideration of an image |
| $Y$ | Total $y$ coordinates of an image |
| $y$ | $y^{th}$ coordinate under consideration of an image |
| $c$ | Channel index |
| $i_c(x,y)$ | $(x,y)$ element of $c^{th}$ channel of an image |
| $L$ | Total number of layers |
| $l$ | Layer number |
| $K_l$ | Total number of kernels of $l^{th}$ layer |
| $k_l$ | Kernel number of $l^{th}$ layer |
| $U$ | Total number of rows of $k^{th}$ kernel |
| $u$ | $u^{th}$ row under consideration |
| $V$ | Total number of columns of $k^{th}$ kernel |
| $v$ | $v^{th}$ column under consideration |
| $e_l^k(u,v)$ | $(u,v)$ element of $k^{th}$ kernel of $l^{th}$ layer |
| $\mathbf{F}_l^k$ | Input feature matrix for $l^{th}$ layer and $k^{th}$ neuron |
| $P$ | Total number of rows of feature matrix |
| $p$ | $p^{th}$ row under consideration |
| $Q$ | Total number of columns of feature matrix |
| $q$ | $q^{th}$ column under consideration |
| $f_l^k(p,q)$ | $(p,q)$ element of feature matrix |
| $g_c(.)$ | Convolution operation |
| $g_p(.)$ | Pooling operation |
| $g_a(.)$ | Activation function |
| $g_k(.)$ | Concatenation operation |
| $g_{t_g}$ | Transformation gate |
| $g_{c_g}$ | Carry gate |
| $g_{sq}(.)$ | Squeeze operation |
| $g_{ex}(.)$ | Excitation operation |
| $\mathbf{Y}_{l+1}^K$ | Weight vector showing feature-maps importance learned using SE operation |
| $g_t$ | Transformation function for two layer NN implemented by SE block |
| $g_{s_g}$ | Sigmoid gate implemented by SE block |
| $g_{sm}$ | Soft mask |
| $g_{tm}$ | Trunk mask |
| $\mathbf{I}_B$ | Channel boosted input tensor |

## 2.2    Pooling layer

Feature motifs, which result as an output of convolution operation, can occur at different locations in the image. Once features are extracted, its exact location becomes less important as long as its approximate position relative to others is preserved. Pooling or down-sampling is an interesting local operation. It sums up similar information in the neighborhood of the receptive field and outputs the dominant response within this local region (Lee et al. 2016).

$$\mathbf{Z}_l^k = g_p(\mathbf{F}_l^k) \tag{2}$$

Equation (2) shows the pooling operation in which $\mathbf{Z}_l^k$ represents the pooled feature-map of $l^{\text{th}}$ layer for $k^{\text{th}}$ input feature-map $\mathbf{F}_l^k$, whereas $g_p(.)$ defines the type of pooling operation.

The use of pooling operation helps to extract a combination of features, which are invariant to translational shifts and small distortions (Ranzato et al. 2007; Scherer et al. 2010). Reduction in the size of feature-map to invariant feature set not only regulates the complexity of the network but also helps in increasing the generalization by reducing overfitting. Different types of pooling formulations such as max, average, L2, overlapping, spatial pyramid pooling, etc. are used in CNN (Boureau 2009; Wang et al. 2012; He et al. 2015b).

## 2.3    Activation function

Activation function serves as a decision function and helps in learning of intricate patterns. The selection of an appropriate activation function can accelerate the learning process. The activation function for a convolved feature-map is defined in equation (3).

$$\mathbf{T}_l^k = g_a(\mathbf{F}_l^k) \tag{3}$$

In the above equation, $\mathbf{F}_l^k$ is an output of a convolution, which is assigned to activation function $g_a(.)$ that adds non-linearity and returns a transformed output $\mathbf{T}_l^k$ for $l^{th}$ layer. In literature, different activation functions such as sigmoid, tanh, maxout, SWISH, ReLU, and variants of ReLU, such as leaky ReLU, ELU, and PReLU are used to inculcate non-linear combination of features (LeCun 2007; Wang et al. 2012; Xu et al. 2015a; Ramachandran et al. 2017; Gu et al. 2018). However, ReLU and its variants are preferred as they help in overcoming the vanishing gradient problem (Hochreiter 1998; Nwankpa et al. 2018). One of the recently proposed activation function is MISH, which has shown better performance than ReLU in most of the recently proposed deep networks on benchmark datasets (Misra 2019).

## 2.4 Batch normalization

Batch normalization is used to address the issues related to the internal covariance shift within feature-maps. The internal covariance shift is a change in the distribution of hidden units' values, which slows down the convergence (by forcing learning rate to small value) and requires careful initialization of parameters. Batch normalization for a transformed feature-map $\mathbf{F}_l^k$ is shown in equation (4).

$$\mathbf{N}_l^k = \frac{\mathbf{F}_l^k - \mu_B}{\sqrt{\sigma_B^2 + \varepsilon}} \tag{4}$$

In equation (4), $\mathbf{N}_l^k$ represents normalized feature-map, $\mathbf{F}_l^k$ is the input feature-map, $\mu_B$ and $\sigma_B^2$ depict mean and variance of a feature-map for a mini batch respectively. In order to avoid division by zero, $\varepsilon$ is added for numerical stability. Batch normalization unifies the distribution of feature-map values by setting them to zero mean and unit variance (Ioffe and Szegedy 2015). Furthermore, it smoothens the flow of gradient and acts as a regulating factor, which thus helps in improving the generalization of the network.

## 2.5 Dropout

Dropout introduces regularization within the network, which ultimately improves generalization by randomly skipping some units or connections with a certain probability. In NNs, multiple connections that learn a non-linear relation are sometimes co-adapted, which causes overfitting (Hinton et al. 2012b). This random dropping of some connections or units produces several thinned network architectures, and finally, one representative network is selected with small weights. This selected architecture is then considered as an approximation of all of the proposed networks (Srivastava et al. 2014).

## 2.6 Fully connected layer

Fully connected layer is mostly used at the end of the network for classification. Unlike pooling and convolution, it is a global operation. It takes input from feature extraction stages and globally analyses the output of all the preceding layers (Lin et al. 2013). Consequently, it makes a non-linear combination of selected features, which are used for the classification of data (Rawat and Wang 2016).

# 3  Architectural evolution of deep CNNs

Nowadays, CNNs are considered as the most widely used algorithms among biologically inspired Artificial Intelligence (AI) techniques. CNN history begins with the neurobiological experiments conducted by Hubel and Wiesel (1959, 1962) (Hubel and Wiesel 1959, 1962). Their work provided a platform for many cognitive models, and CNN replaced almost all of these. Over the decades, different efforts have been carried out to improve the performance of CNNs. The evolutionary history of deep CNN architectures is pictorially represented in Fig. 3. Improvements in CNN architectures can be categorized into five different eras that are discussed below.

## 3.1    Origin of CNN: Late 1980s-1999

CNNs have been applied to visual tasks since the late 1980s. In 1989, LeCuN et al. proposed the first multilayered CNN named ConvNet, whose origin rooted in Fukushima's Neocognitron (Fukushima and Miyake 1982; Fukushima 1988). LeCuN proposed a supervised training of ConvNet using the backpropagation algorithm, in comparison to the unsupervised reinforcement learning scheme used by its predecessor Neocognitron (Linnainmaa 1970; LeCun et al. 1989). LeCuN's work thus made a foundation for the modern 2D CNNs. This ConvNet showed successful results for handwritten digit and zip code recognition related problems (Zhang and LeCun 2015). In 1998, LeCuN proposed an improved version of ConvNet, which was famously known as LeNet-5, and it started the use of CNN in classifying characters in a document recognition related applications (LeCun et al. 1995, 1998). Due to the good performance of CNN in optical character and fingerprint recognition, its commercial use in ATM and Banks started in 1993 and 1996, respectively. In this era, LeNet-5 achieved many successful milestones for optical character recognition tasks, but it didn't perform well on other image recognition problems.

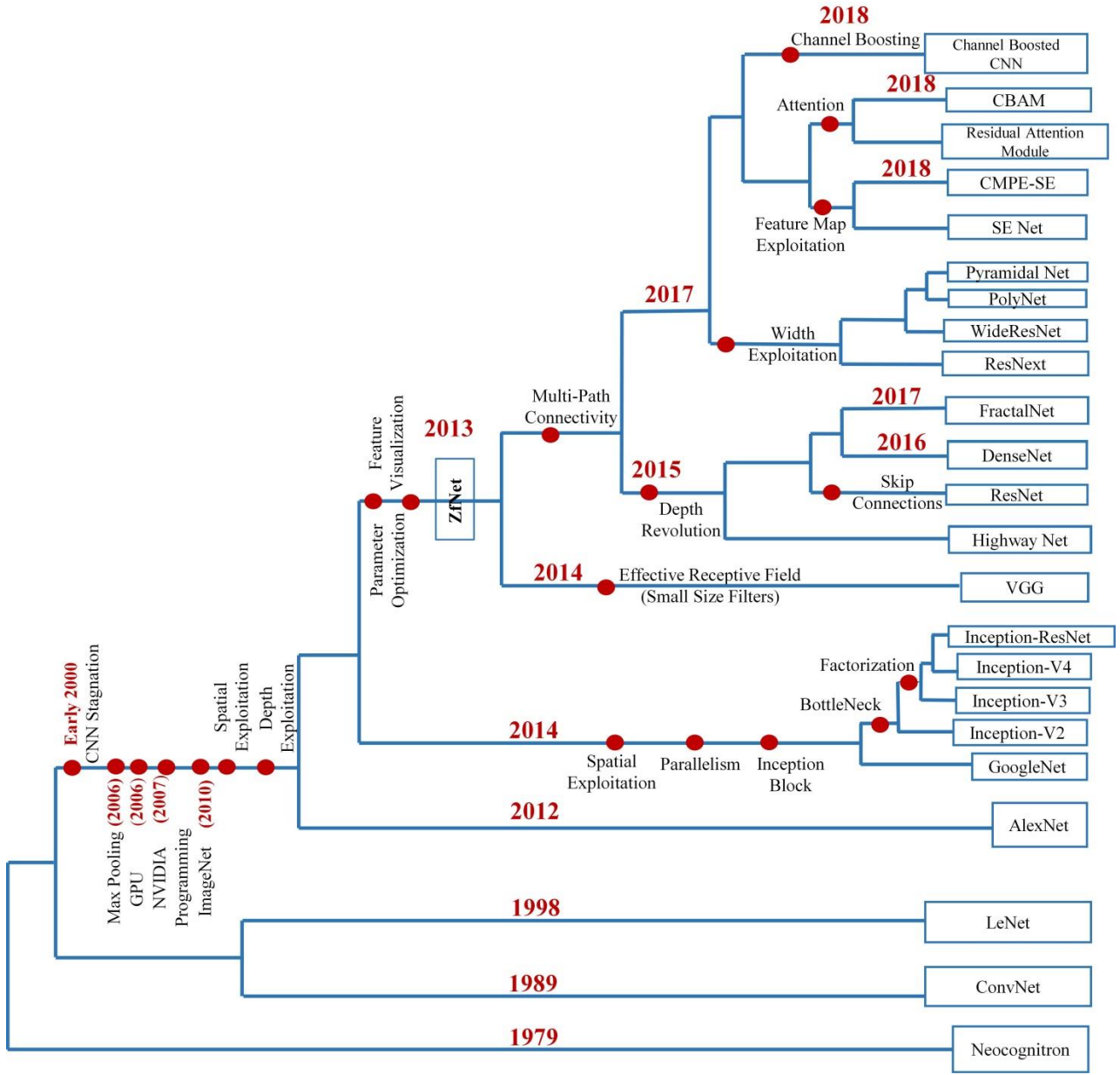

**Fig. 3** Evolutionary history of deep CNNs showing architectural innovations from ConvNet till to date architectures.

## 3.2    Stagnation of CNN: Early 2000

In the late 1990s and early 2000, researchers had little insight into the internal working of CNN, and it was considered as a black box. Complicated architecture design and heavy processing made it hard to train CNN. It was widely presumed in early 2000 that the backpropagation algorithm used for training of CNN was not effective in converging to the global minima of the error surface. Thus, CNN was considered as a less effective feature extractor compared to handcrafted features (Schmidhuber 2007). Moreover, no comprehensive dataset of diverse categories of images was available at that time. Therefore, because of the insignificant improvement in CNN performance at the cost of high computational time, little attention was given to explore its role in different applications such as object detection, video surveillance, etc. At that time, other statistical methods and in particular, SVM became more popular than CNN due to their relatively high performance (Joachims 1998; Decoste and Schölkopf 2002; Liu et al. 2003).

Meanwhile, a few research groups kept on working with CNNs and tried to optimize its performance. In 2003, Simard et al. improved CNN architecture and showed good results compared to SVM on a hand digit benchmark dataset; MNIST (LeCun et al. 1998; Liu et al. 2003; Simard et al. 2003; Chellapilla et al. 2006; Deng 2012). This improvement in performance expedited the research in CNNs by extending their application's beyond optical character recognition to other script's character recognition, deployment in image sensors for face detection in video conferencing, and regulation of street crimes, etc. (Abdulkader 2006; Chellapilla et al. 2006; Ciresan et al. 2010). Likewise, CNN based systems were industrialized in markets for customers' tracking (Garcia and Delakis 2004; Frome et al. 2009; LeCun et al. 2010). Moreover, CNN's potential in other applications such as medical image segmentation, anomaly detection, and robot vision was also explored (Fasel 2002; Matsugu et al. 2002; Chen et al. 2006).

## 3.3    Revival of CNN: 2006-2011

Deep CNNs generally have complex architecture and time-intensive training phase that sometimes may span over weeks. In early 2000, there were a few parallel processing techniques and limited hardware resources for the training of deep Networks. Training of a deep CNNs with a typical activation function such as sigmoid may suffer from exponential decay and explosion of a gradient. Since 2006, significant efforts have been made to tackle the CNN optimization problem. In this regard, several interesting initialization and training strategies were reported to overcome the difficulties encountered in the training of deep CNNs and the learning of invariant features. Hinton reported the concept of greedy layer-wise pre-training in 2006, which revived the research in deep learning (Hinton et al. 2006; Khan et al. 2018b). Experimental studies showed that both supervised and unsupervised pre-training could initialize a network in a better way than random initialization. Bengio and other researchers proposed that the sigmoid activation function is not suitable for the training of deep architectures with random initialization of weights. This observation started the use of activation functions other than sigmoid such as ReLU, tanh etc., (Glorot and Bengio 2010). The revival of deep learning was one of the factors, which brought deep CNNs into limelight (Bengio et al. 2007, 2013).

Ranzato et al. (2007) used max-pooling instead of subsampling, which showed good results by learning invariant features (Ranzato et al. 2007; Giusti et al. 2013). In late 2006, researchers started using graphics processing units (GPUs) to accelerate the training of deep NN and CNN architectures (Oh and Jung 2004; Strigl et al. 2010; Cireşan et al. 2011; Nguyen et al. 2019). In 2007, NVIDIA launched the CUDA programming platform, which allows exploitation of parallel processing capabilities of GPU with a greater degree (Nickolls et al. 2008; Lindholm et al. 2008). In essence, the use of GPUs for NN and CNN training and other hardware improvements were the main factors, which revived the research in CNN (Oh and Jung 2004; Ciresan et al. 2018). In 2010, Fei-Fei Li's group at Stanford, established a large database of images known as ImageNet, containing millions of annotated images belonging to a large number of classes (Russakovsky et al. 2015). This database was coupled with the annual ImageNet Large Scale Visual Recognition Challenge (ILSVRC), where the performances of various models have been evaluated and scored (Berg et al. 2010). Similarly, in the same year, Stanford released PASCAL 2010 VOC dataset for object detection. ILSVRC and Neural Information Processing Systems Conference (NIPS) are the two platforms that play a dominant role in strengthening research and increasing the use of CNN and thus making it popular.

### 3.4    Rise of CNN: 2012-2014

The availability of extensive training data and hardware advancements are the factors that contributed to the advancement in CNN research. But the main driving forces that have accelerated the research and give rise to the use of CNNs in image classification and recognition tasks are parameter optimization strategies and new architectural ideas (Gu et al. 2018; Sinha et al. 2018; Zhang et al. 2019). The main breakthrough in CNN performance was brought by AlexNet, which showed exemplary performance in 2012-ILSVRC (reduced error rate from 25.8 to 16.4) as compared to conventional CV techniques (Krizhevsky et al. 2012).

In this era, several attempts were made to improve the performance of CNN; depth and parameter optimization strategies were explored with a significant reduction in computational cost. Similarly, different architectural designs were proposed, whereby each new architecture tried to overcome the shortcomings of previously proposed architectures in combination with new structural reformulations. With the trend of designing very deep CNNs, it generally becomes difficult to independently determine filter dimensions, stride, padding, and other hyper-parameters for each layer. This problem is resolved by designing convolutional layers with a fixed topology that can be repeated multiple times. This shifted the trend from custom layer design towards modular and uniform layer design. The concept of modularity in CNNs made it easy to tailor them for different tasks effortlessly (Simonyan and Zisserman 2015; Amer and Maul 2019). In this connection, a different idea of branching and block within a layer was introduced by the Google group (Szegedy et al. 2015). It should be noted that in this era, two different types of architectures, deep and narrow, as well as deep and wide, were in use.

## 3.5 Rapid increase in architectural innovations and applications of CNN: 2015-Present

The research in CNN is still going on and has a significant potential for improvement. It is generally observed that the significant improvements in CNN performance occurred from 2015-2019. The representational capacity of a CNN usually depends on its depth, and in a sense, an enriched feature set ranging from simple to complex abstractions can help in learning complex problems. However, the main challenge faced by deep architectures is that of the diminishing gradient. Initially, researchers tried to subside this problem by connecting intermediate layers to auxiliary learners (Szegedy et al. 2015). In 2015, the emerging area of research was mainly the development of new connections to improve the convergence rate of deep CNN architectures. In this regard, different ideas such as information gating mechanism across multiple layers, skip connections, and cross-layer channel connectivity was introduced (Srivastava et al. 2015a; He et al. 2015a; Huang et al. 2017). Different experimental studies showed that state-of-the-art deep architectures such as VGG, ResNet, ResNext, etc. also showed good results for challenging recognition and localization problems like semantic and instance-based object segmentation, scene parsing, scene location, etc. Most of the famous object detection and segmentation architectures such as Single Shot Multibox Detector (SSD), Region-based CNN (R-CNN), Faster R-CNN, Mask R-CNN and Fully Convolutional Neural Network (FCN) are built on the lines of ResNet, VGG, Inception, etc. Similarly, many interesting detection algorithms such as Feature Pyramid Networks, Cascade R-CNN, Libra R-CNN, etc., modified the architectures as mentioned earlier to improve the performance (Lin et al. 2017; Cai and Vasconcelos 2019; Pang et al. 2020). Applications of deep CNN were also extended to image captioning by combining these networks with recurrent neural network (RNN) and thus showed state-of-the-art results on MS COCO-2015 image captioning challenge (Girshick 2015; Long et al. 2015; Ren et al. 2015; He et al. 2017; Vinyals et al. 2017).

Similarly, in 2016, it was observed that the stacking of multiple transformations not only depth-wise but also in parallel fashion showed good learning for complex problems (Zagoruyko and Komodakis 2016; Han et al. 2017). Different researchers used a hybrid of the already proposed architectures to improve deep CNN performance (Huang et al. 2016a; Szegedy et al. 2016a; Targ et al. 2016; Yamada et al. 2016; Kuen et al. 2017; Lv et al. 2019). In 2017, the focus of researchers was mainly on designing of generic blocks that can be inserted at any learning stage in CNN architecture to improve the network representation (Hu et al. 2018a). Designing of new blocks is one of the growing areas of research in CNN, where generic blocks are used to assign attention to spatial and feature-map (channel) information (Wang et al. 2017a; Roy et al. 2018; Woo et al. 2018). In 2018, a new idea of channel boosting was introduced by Khan et al. (Khan et al. 2018a) to boost the performance of a CNN by learning distinct features as well as exploiting the already learned features through the concept of TL.

However, two main concerns observed with deep and wide architectures are the high computational cost and memory requirement. As a result, it is very challenging to deploy state-of-the-art wide and deep CNN models in resource-constrained environments. Conventional convolution operation requires a huge number of multiplications, which increases the inference time and restricts the applicability of CNN to low memory and time constraint applications (Shakeel et al. 2019). Many real-world applications, such as autonomous vehicles, robotics, healthcare, and mobile applications, perform the tasks that need to be carried on computationally limited platforms in a timely manner. Therefore, different modifications in CNN are performed to make them appropriate for resource-constrained environments. Prominent modifications are knowledge distillation, training of small networks, or squeezing of pre-trained networks (such as pruning, quantization, hashing, Huffman coding, etc.) (Chen et al. 2015; Han et al. 2016; Wu et al. 2016; Frosst and Hinton 2018). GoogleNet exploited the idea of small networks, which replaces the conventional convolution with point-wise group convolution operation to make it computationally efficient. Similarly, ShuffleNet used point-wise group convolution but with a new idea of channel shuffle that significantly reduces the number of operations without affecting the accuracy. In the same way, ANTNet proposed a novel architectural block known as ANTBlock, which at low computational cost, achieved good performance on benchmark datasets (Howard et al. 2017; Zhang et al. 2018a; Xiong et al. 2019).

From 2012 up till now, many improvements have been reported in CNN architectures. As regards the architectural advancement of CNNs, recently, the focus of research has been on designing of new blocks that can boost network representation by exploiting feature-maps or manipulating input representation by adding artificial channels. Moreover, along with this, the trend is towards the design of lightweight architectures without compromising the performance to make CNN applicable for resource constraint hardware.

## 4   Architectural innovations in CNN

Different improvements in CNN architecture have been made from 1989 to date. These improvements can be categorized as parameter optimization, regularization, structural reformulation, etc. However, it is observed that the main thrust in CNN performance improvement came from the restructuring of processing units and the designing of new blocks. Most of the innovations in CNN architectures have been made in relation to depth and spatial exploitation. Depending upon the type of architectural modifications, CNNs can be broadly categorized into seven different classes, namely; spatial exploitation, depth, multi-path, width, feature-map exploitation, channel boosting, and attention-based CNNs. The taxonomy of CNN architectures is pictorially represented in Fig. 4. Architectural details of the state-of-the-art CNN models, their parameters, and performance on benchmark datasets are summarized in Table 2. On the other hand, different online resources on deep CNN architectures, vision-related dataset, and their implementation platforms are mentioned in Table 3. In addition to this, the strengths and weaknesses of various architectures based on their category are presented in Table 5a-g.

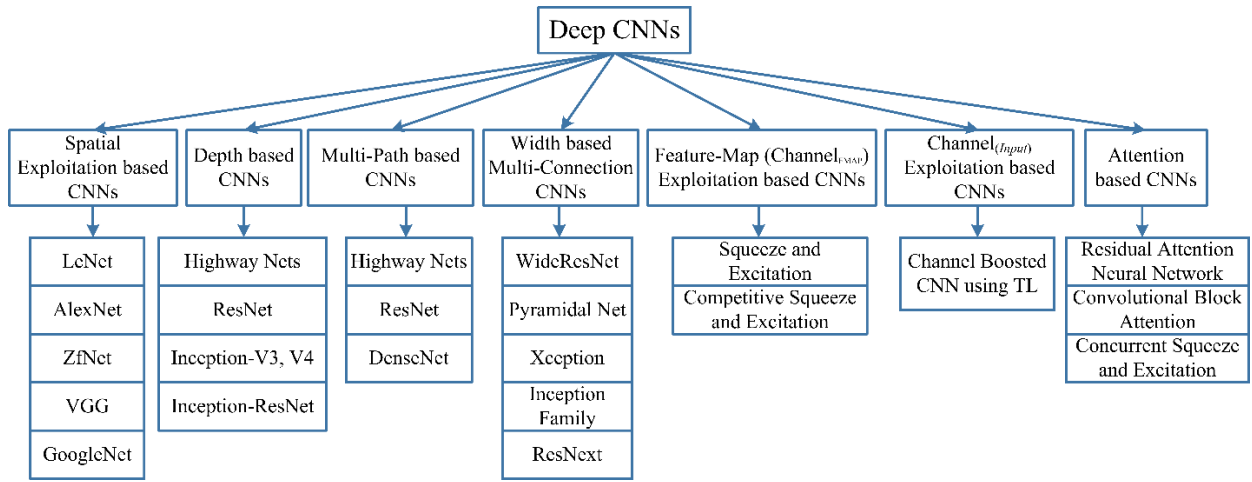

**Fig. 4** Taxonomy of deep CNN architectures showing seven different categories.

## 4.1    Spatial Exploitation based CNNs

CNNs have a large number of parameters and hyper-parameters, such as weights, biases, number of layers, and processing units (neurons), filter size, stride, activation function, learning rate, etc. (Kafi et al. 2015; Shin et al. 2016). As convolutional operation considers the neighborhood (locality) of input pixels, therefore different levels of correlation can be explored by using different filter sizes. Different sizes of filters encapsulate different levels of granularity; usually, small size filters extract fine-grained and large size extract coarse-grained information. Consequently, in early 2000, researchers exploited spatial filters to improve performance and explored the relation of a spatial filter with the learning of the network. Different studies conducted in this era suggested that by the adjustment of filters, CNN can perform well both on coarse and fine-grained details.

### 4.1.1    LeNet

LeNet was proposed by LeCuN in 1998 (LeCun et al. 1995). It is famous due to its historical importance as it was the first CNN, which showed state-of-the-art performance on hand digit recognition tasks. It has the ability to classify digits without being affected by small distortions,

**Table 2** Performance comparison of the recent architectures of different categories. Top 5 error rate is reported for all architectures.

| Architecture Name | Year | Main contribution | Parameters | Error Rate | Depth | Category | Reference |
|---|---|---|---|---|---|---|---|
| LeNet | 1998 | - First popular CNN architecture | 0.060 M | [dist]MNIST: 0.8 MNIST: 0.95 | 5 | Spatial Exploitation | (LeCun et al. 1995) |
| AlexNet | 2012 | - Deeper and wider than the LeNet - Uses Relu, dropout and overlap Pooling - GPUs NVIDIA GTX 580 | 60 M | ImageNet: 16.4 | 8 | Spatial Exploitation | (Krizhevsky et al. 2012) |
| ZfNet | 2014 | -Visualization of intermediate layers | 60 M | ImageNet: 11.7 | 8 | Spatial Exploitation | (Zeiler and Fergus 2013) |
| VGG | 2014 | - Homogenous topology - Uses small size kernels | 138 M | ImageNet: 7.3 | 19 | Spatial Exploitation | (Simonyan and Zisserman 2015) |
| GoogLeNet | 2015 | - Introduced block concept - Split transform and merge idea | 4 M | ImageNet: 6.7 | 22 | Spatial Exploitation | (Szegedy et al. 2015) |
| Inception-V3 | 2015 | - Handles the problem of a representational bottleneck - Replace large size filters with small filters | 23.6 M | ImageNet: 3.5 Multi-Crop: 3.58 Single-Crop: 5.6 | 159 | Depth + Width | (Szegedy et al. 2016b) |
| Highway Networks | 2015 | - Introduced an idea of Multi-path | 2.3 M | CIFAR-10: 7.76 | 19 | Depth + Multi-Path | (Srivastava et al. 2015a) |
| Inception-V4 | 2016 | - Split transform and merge idea Uses asymmetric filters | 35 M | ImageNet: 4.01 | 70 | Depth +Width | (Szegedy et al. 2016a) |
| Inception-ResNet | 2016 | - Uses split transform merge idea and residual links | 55.8M | ImageNet: 3.52 | 572 | Depth + Width + Multi-Path | (Szegedy et al. 2016a) |
| ResNet | 2016 | - Residual learning - Identity mapping based skip connections | 25.6 M 1.7 M | ImageNet: 3.6 CIFAR-10: 6.43 | 152 110 | Depth + Multi-Path | (He et al. 2015a) |
| DelugeNet | 2016 | - Allows cross layer information flow in deep networks | 20.2 M | CIFAR-10: 3.76 CIFAR-100: 19.02 | 146 | Multi-path | (Kuen et al. 2018) |
| FractalNet | 2016 | - Different path lengths are interacting with each other without any residual connection | 38.6 M | CIFAR-10: 7.27 CIFAR-10+: 4.60 CIFAR-10++: 4.59 CIFAR-100: 28.20 CIFAR-100+: 22.49 CIFAR100++: 21.49 | 20 40 | Multi-Path | (Larsson et al. 2016) |
| WideResNet | 2016 | - Width is increased and depth is decreased | 36.5 M | CIFAR-10: 3.89 CIFAR-100: 18.85 | 28 - | Width | (Zagoruyko and Komodakis 2016) |
| Xception | 2017 | - Depth wise convolution followed by point wise convolution | 22.8 M | ImageNet: 0.055 | 126 | Width | (Chollet 2017) |
| Residual Attention Neural Network | 2017 | - Introduced an attention mechanism | 8.6 M | CIFAR-10: 3.90 CIFAR-100: 20.4 ImageNet: 4.8 | 452 | Attention | (Wang et al. 2017a) |
| ResNeXt | 2017 | - Cardinality - Homogeneous topology - Grouped convolution | 68.1 M | CIFAR-10: 3.58 CIFAR-100: 17.31 ImageNet: 4.4 | 29 - 101 | Width | (Xie et al. 2017) |
| Squeeze & Excitation Networks | 2017 | - Models interdependencies between feature-maps | 27.5 M | ImageNet: 2.3 | 152 | Feature-Map Exploitation | (Hu et al. 2018a) |
| DenseNet | 2017 | - Cross-layer information flow | 25.6 M 25.6 M 15.3 M 15.3 M | CIFAR-10+: 3.46 CIFAR100+:17.18 CIFAR-10: 5.19 CIFAR-100: 19.64 | 190 190 250 250 | Multi-Path | (Huang et al. 2017) |
| PolyNet | 2017 | - Experimented structural diversity - Introduced Poly Inception module - Generalizes residual unit using polynomial compositions | 92 M | ImageNet: Single:4.25 Multi:3.45 | - - | Width | (Zhang et al. 2017) |
| PyramidalNet | 2017 | - Increases width gradually per unit | 116.4 M 27.0 M 27.0 M | ImageNet: 4.7 CIFAR-10: 3.48 CIFAR-100: 17.01 | 200 164 164 | Width | (Han et al. 2017) |
| Convolutional Block Attention Module (ResNeXt101 (32x4d) + CBAM) | 2018 | - Exploits both spatial and feature-map information | 48.96 M | ImageNet: 5.59 | 101 | Attention | (Woo et al. 2018) |
| Concurrent Spatial & Channel Excitation Mechanism | 2018 | - Spatial attention - Feature-map attention - Concurrent placement of spatial and channel attention | - | MALC: 0.12 Visceral: 0.09 | - | Attention | (Roy et al. 2018) |
| Channel Boosted CNN | 2018 | - Boosting of original channels with additional information rich generated artificial channels | - | - | - | Channel Boosting | (Khan et al. 2018a) |
| Competitive Squeeze & Excitation Network CMPE-SE-WRN-28 | 2018 | - Residual and identity mappings both are used for rescaling the feature-map | 36.92 M 36.90 M | CIFAR-10: 3.58 CIFAR-100: 18.47 | 152 152 | Feature-Map Exploitation | (Hu et al. 2018b) |

**Table 3** Different online available resources for deep CNNs.

| Category | Description | Source |
|---|---|---|
| Cloud based Platforms | Online free access to GPU and other deep learning accelerators | Google Colab: https://colab.research.google.com/notebooks/welcome.ipynb |
| | | Colac: https://cocalc.com/ |
| | Commercial platforms offered by world leading companies | FloydHub: https://www.floydhub.com/ |
| | | Amazon SageMaker: https://aws.amazon.com/deep-learning/ |
| | | Microsoft Azure ML Services: https://azure.microsoft.com/en-gb/services/machine-learning/ |
| | | Google Cloud: https://cloud.google.com/deep-learning-vm/ |
| | | IBM Watson Studio: https://www.ibm.com/cloud/deep-learning |
| Deep Learning Libraries | Deep learning libraries that provide built-in classes of NNs, fast numerical computation and automated estimation of gradients both for CPU and GPU | Pytorch: https://pytorch.org/ |
| | | Tensorflow: https://www.tensorflow.org/ |
| | | MatConvNet: http://www.vlfeat.org/matconvnet/ |
| | | Keras: https://keras.io/ |
| | | Theano: http://deeplearning.net/software/theano/ |
| | | Caffe: https://caffe.berkeleyvision.org/ |
| | | Julia: https://julialang.org/ |
| Lecture Series | Online available and freely accessible deep learning courses | Stanford Lecture Series: http://cs231n.stanford.edu/ |
| | | Udacity: https://www.udacity.com/course/deep-learning-nanodegree--nd101 https://www.udacity.com/course/deep-learning-pytorch--ud188 |
| | | Udemy: https://www.udemy.com/course/deep-learning-learn-cnns/ https://www.udemy.com/course/modern-deep-convolutional-neural-networks/ https://www.udemy.com/course/advanced-computer-vision/ https://www.udemy.com/course/deep-learning-convolutional-neural-networks-theano-tensorflow/ https://www.udemy.com/course/deep-learning-pytorch/ |
| | | Coursera: https://www.coursera.org/learn/convolutional-neural-networks https://www.coursera.org/specializations/deep-learning |
| Vision Datasets | Online freely accessible datasets of different categories of annotated images | ImageNet: http://image-net.org/ |
| | | COCO: http://cocodataset.org/#home |
| | | Visual Genome: http://visualgenome.org/ |
| | | Open images: https://ai.googleblog.com/2016/09/introducing-open-images-dataset.html |
| | | Places: http://places.csail.mit.edu/index.html |
| | | Youtube-8M: https://research.google.com/youtube8m/index.html |
| | | CelebA: http://mmlab.ie.cuhk.edu.hk/projects/CelebA.html |
| | | CIFAR10: https://www.cs.toronto.edu/~kriz/cifar.html |
| | | Indoor Scene Recognition: http://web.mit.edu/torralba/www/indoor.html |
| | | Computer Vision Datasets: https://computervisiononline.com/datasets |
| | | Fashion MNIST: https://research.zalando.com/welcome/mission/research-projects/fashion-mnist/ |
| Deep Learning Accelerators | Energy and computation efficient deep learning accelerators | NVIDIA: http://nvdla.org/ |
| | | FPGA: https://www.intel.com/content/www/us/en/artificial-intelligence/programmable/overview.html |
| | | Eyeriss: http://eyeriss.mit.edu/ |
| | | Google's TPU: https://cloud.google.com/tpu/ |

rotation, and variation of position and scale. LeNet is a feed-forward NN that constitutes of five alternating layers of convolutional and pooling, followed by two fully connected layers. In early 2000, GPU was not commonly used to speed up training, and even CPUs were slow (Potluri et al. 2011). The main limitation of traditional multilayered fully connected NN was that it considers each pixel as separate input and applies a transformation on it, which was a substantial computational burden, specifically at that time (Gardner and Dorling 1998). LeNet exploited the underlying basis of the image that the neighboring pixels are correlated to each other and feature motifs are distributed across the entire image. Therefore, convolution with learnable parameters is an effective way to extract similar features at multiple locations with few parameters. Learning with sharable parameters changed the conventional view of training where each pixel was considered as a separate input feature from its neighborhood and ignored the correlation among them. LeNet was the first CNN architecture, which not only reduced the number of parameters but was able to learn features from raw pixels automatically.

### 4.1.2   AlexNet

LeNet (LeCun et al. 1995) though, begin the history of deep CNNs, but at that time, CNN was limited to hand digit recognition tasks and didn't perform well to all classes of images. AlexNet (Krizhevsky et al. 2012) is considered as the first deep CNN architecture, which showed groundbreaking results for image classification and recognition tasks. AlexNet was proposed by Krizhevesky et al., which enhanced the learning capacity of the CNN by making it deeper and by applying several parameter optimizations strategies (Krizhevsky et al. 2012). The basic architectural design of AlexNet is shown in Fig. 5. In early 2000, hardware limitations curtailed the learning capacity of deep CNN architectures by restricting them to small size. In order to get the benefit of the representational capacity of deep CNNs, Alexnet was trained in parallel on two NVIDIA GTX 580 GPUs to overcome shortcomings of the hardware.

In AlexNet, depth was extended from 5 (LeNet) to 8 layers to make CNN applicable for diverse categories of images. Despite the fact that generally, depth improves generalization for different resolutions of images but, the main drawback associated with an increase in depth is overfitting. To address this challenge, Krizhevesky et al. (2012) exploited the idea of Hinton (Dahl et al. 2013; Srivastava et al. 2014), whereby their algorithm randomly skips some transformational units during training to enforce the model to learn more robust features. In addition to this, ReLU was employed as a non-saturating activation function to improve the convergence rate by alleviating the problem of vanishing gradient to some extent (Hochreiter 1998; Nair and Hinton 2010). Overlapping subsampling and local response normalization were also applied to improve the generalization by reducing overfitting. Other adjustments made were the use of large size filters (11x11 and 5x5) at the initial layers, compared to previously proposed networks. Due to the efficient learning approach of AlexNet, it has significant importance in the new generation of CNNs and has started a new era of research in the architectural advancements of CNNs.

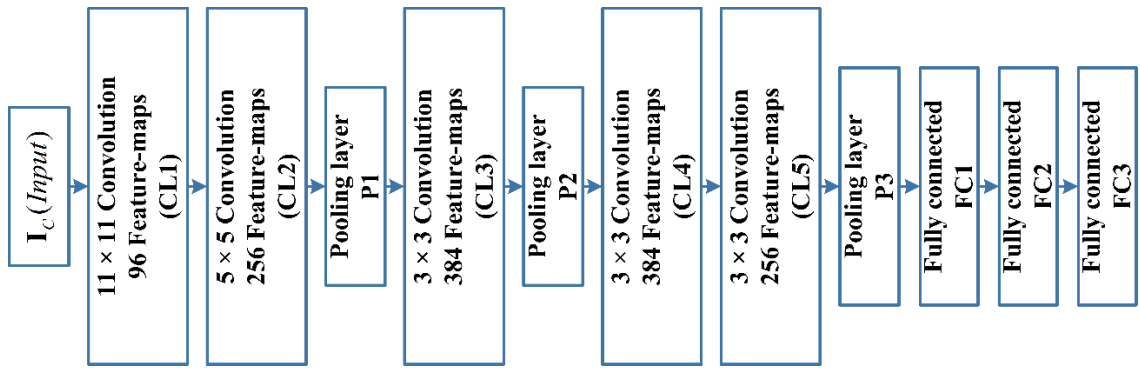

**Fig. 5** Basic layout of AlexNet architecture showing its five convolution and three fully connected layers.

### 4.1.3 ZfNet

The learning mechanism of CNN, before 2013, was based mainly on hit-and-trial, without knowing the exact reason behind the improvement. This lack of understanding limited the performance of deep CNNs on complex images. In 2013, Zeiler and Fergus proposed an interesting multilayer Deconvolutional NN (DeconvNet), which got famous as ZfNet (Zeiler and Fergus 2013). ZfNet was developed to visualize network performance quantitatively. The idea of the visualization of network activity was to monitor CNN performance by interpreting neuron's activation. In one of the previous studies, Erhan et al. (2009) exploited the same idea and optimized the performance of Deep Belief Networks (DBNs) by visualizing the hidden layers' feature (Erhan et al. 2009). Similarly, Le et al. (2011) evaluated the learning of deep unsupervised autoencoder (AE) by visualizing the image classes generated by the neurons of last layer (Le et al. 2011). DeconvNet works in the same manner as the forward pass CNN but reverses the order of convolutional and pooling operation. This reverse mapping projects the output of the convolutional layer back to visually perceptible image patterns, consequently gives the neuron-level interpretation of the internal feature representation learned at each layer (Simonyan et al. 2013; Grün et al. 2016).

The idea of feature visualization proposed by ZfNet was experimentally validated on AlexNet using DeconvNet, which showed that only a few neurons were active. In contrast, other neurons were dead (inactive) in the first and second layers of the network. Moreover, it showed that the features extracted by the second layer exhibited aliasing artifacts. Based on these findings, Zeiler and Fergus adjusted CNN topology and performed parameter optimization. Zeiler and Fergus maximized the learning of CNN by reducing both the filter size and stride to retain the maximum number of features in the first two convolutional layers. This readjustment in CNN topology resulted in performance improvement, which suggested that features visualization can be used for the identification of design shortcomings and for timely adjustment of parameters.

### 4.1.4 VGG

The successful use of CNNs in image recognition tasks has accelerated the research in architectural design. In this regard, Simonyan et al. proposed a simple and effective design principle for CNN architectures. Their architecture, named as VGG, was modular in layers pattern (Simonyan and Zisserman 2015). VGG was made 19 layers deep compared to AlexNet and ZfNet to simulate the relation of depth with the representational capacity of the network (Krizhevsky et al. 2012; Zeiler and Fergus 2013). ZfNet, which was a frontline network of 2013-ILSVRC competition, suggested that small size filters can improve the performance of the CNNs. Based on these findings, VGG replaced the 11x11 and 5x5 filters with a stack of 3x3 filters layer and experimentally demonstrated that concurrent placement of small size (3x3) filters could induce the effect of the large size filter (5x5 and 7x7). The use of the small size filters provides an additional benefit of low computational complexity by reducing the number of parameters. These findings set a new trend in research to work with smaller size filters in CNN. VGG regulates the complexity of a network by placing 1x1 convolutions in between the convolutional layers, which, besides, learn a linear combination of the resultant feature-maps.

For the tuning of the network, max-pooling is placed after the convolutional layer, while padding was performed to maintain the spatial resolution (Ranzato et al. 2007). VGG showed good results both for image classification and localization problems. VGG was at $2^{nd}$ place in the 2014-ILSVRC competition but, got fame due to its simplicity, homogenous topology, and increased depth. The main limitation associated with VGG was the use of 138 million parameters, which make it computationally expensive and difficult to deploy it on low resource systems.

### 4.1.5 GoogleNet

GoogleNet was the winner of the 2014-ILSVRC competition and is also known as Inception-V1. The main objective of the GoogleNet architecture was to achieve high accuracy with a reduced computational cost (Szegedy et al. 2015). It introduced the new concept of inception block in CNN, whereby it incorporates multi-scale convolutional transformations using split, transform and merge idea. The architecture of the inception block is shown in Fig. 6. In GoogleNet, conventional convolutional layers are replaced in small blocks similar to the idea of substituting each layer with micro NN as proposed in Network in Network (NIN) architecture (Lin et al. 2013). This block encapsulates filters of different sizes (1x1, 3x3, and 5x5) to capture spatial information at different scales, including both fine and coarse grain level. The exploitation of the idea of split, transform, and merge by GoogleNet, helped in addressing a problem related to the learning of diverse types of variations present in the same category of images having different resolutions. GoogleNet regulates the computations by adding a bottleneck layer of 1x1 convolutional filter, before employing large size kernels. In addition to it, it used sparse connections (not all the output feature-maps are connected to all the input feature-maps), to overcome the problem of redundant information and reduced cost by omitting feature-maps that were not relevant. Furthermore, connection's density was reduced by using global average pooling at the last layer, instead of using a fully connected layer. These parameter tunings caused a significant decrease in the number of parameters from 138 million to 4 million parameters. Other regulatory factors applied were batch normalization and the use of RmsProp as an optimizer (Dauphin et al. 2015). GoogleNet also introduced the concept of auxiliary learners to speed up the convergence rate. However, the main drawback of the GoogleNet was its heterogeneous topology that needs to be customized from module to module. Another limitation of GoogleNet was a representation bottleneck that drastically reduces the feature space in the next layer and thus sometimes may lead to loss of useful information.

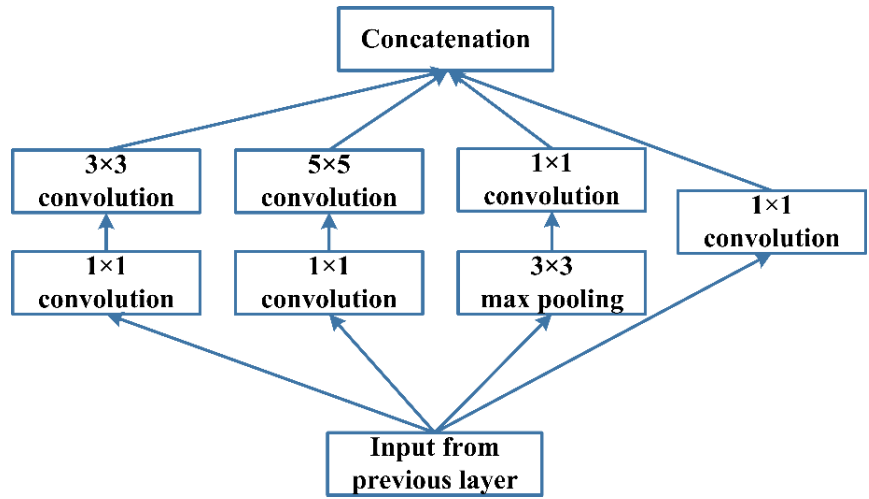

**Fig. 6** Basic architecture of the inception block showing the split, transform, and merge concept.

## 4.2 Depth based CNNs

Deep CNN architectures are based on the assumption that with the increase in depth, the network can better approximate the target function with a number of nonlinear mappings and more enriched feature hierarchies (Bengio 2013). Network depth has played an essential role in the success of supervised training. Theoretical studies have shown that deep networks can represent certain classes of function more efficiently than shallow architectures (Montufar et al. 2014). Csáji represented a universal approximation theorem in 2001, which states that a single hidden layer is sufficient to approximate any function. However, this comes at the cost of exponentially many neurons; thus, it often makes it computationally non-realistic (Csáji 2001). In this regard, Bengio and Delalleau (Delalleau and Bengio 2011) suggested that deeper networks can maintain the expressive power of the network at a reduced cost (Wang and Raj 2017). In 2013, Bengio et al. empirically showed that deep networks are computationally more efficient for complex tasks (Bengio et al. 2013; Nguyen et al. 2018). Inception and VGG, which showed the best performance in 2014-ILSVRC competition, further strengthen the idea that the depth is an essential dimension in regulating learning capacity of the networks (Simonyan and Zisserman 2015; Szegedy et al. 2015, 2016a, b).

### 4.2.1 Highway Networks

Based on the intuition that the learning capacity can be improved by increasing the network depth, Srivastava et al. in 2015, proposed a deep CNN, named as Highway Networks (Srivastava et al. 2015a). The main problem concerned with deep networks is slow training and convergence speed (Huang et al. 2016b). Highway Networks exploited depth for learning enriched feature representation and introducing a new cross-layer connectivity mechanism (discussed in Section 4.3.1) for the successful training of the deep networks. Therefore, Highway Networks are also categorized as multi-path based CNN architectures. Highway Networks with 50-layers showed a better convergence rate than thin but deep architectures (Berg et al. 2010; Morar et al. 2012). Srivastava et al. experimentally showed that the performance of a plain network decreases after adding hidden units beyond 10 layers (Glorot and Bengio 2010). Highway Networks, on the other hand, was shown to converge significantly faster than the plain ones, even with the depth of 900 layers.

### 4.2.2 ResNet

ResNet was proposed by He et al., which is considered as a continuation of deep networks (He et al. 2015a). ResNet revolutionized the CNN architectural race by introducing the concept of residual learning in CNNs and devised an efficient methodology for the training of deep networks. Similar to Highway Networks, it is also placed under the Multi-Path based CNNs; thus, its learning methodology is discussed in Section 4.3.2. ResNet proposed 152-layers deep CNN, which won the 2015-ILSVRC competition. The architecture of the residual block of ResNet is shown in Fig. 7. ResNet, which was 20 and 8 times deeper than AlexNet and VGG, respectively, showed less computational complexity than previously proposed networks (Krizhevsky et al. 2012; Simonyan and Zisserman 2015). He et al. empirically showed that ResNet with 50/101/152 layers has less error on image classification task than 34 layers plain Net. Moreover, ResNet gained a 28% improvement on the famous image recognition benchmark dataset named COCO (Lin et al. 2014). Good performance of ResNet on image recognition and localization tasks showed that representational depth is of central importance for many visual recognition tasks.

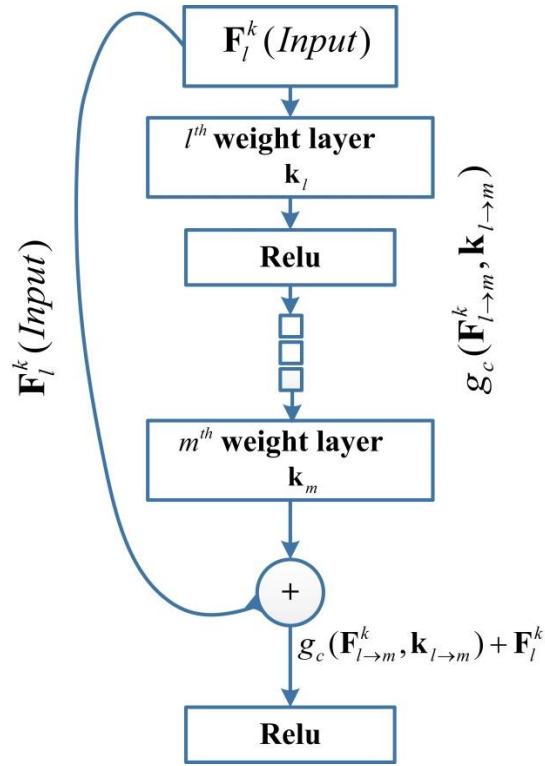

**Fig. 7** Residual block as a basic structural unit of ResNet.

### 4.2.3    Inception-V3, V4 and Inception-ResNet

Inception-V3, V4 and Inception-ResNet, are improved versions of Inception-V1 and V2 (Szegedy et al. 2015, 2016a, b). The idea of Inception-V3 was to reduce the computational cost of deep networks without affecting the generalization. For this purpose, Szegedy et al. replaced large size filters (5x5 and 7x7) with small and asymmetric filters (1x7 and 1x5) and used 1x1 convolution as a bottleneck before the large filters (Szegedy et al. 2016b). Concurrent placement of 1x1 convolution with a large size filter makes the traditional convolution operation more like a cross-channel correlation. In one of the previous works, Lin et al. exploited the potential of 1x1 filters in NIN architecture (Lin et al. 2013). Szegedy et al. (Szegedy et al. 2016b) intelligently used the same concept. In Inception-V3, 1x1 convolutional operation was used, which maps the input data into 3 or 4 separate spaces that are smaller than the original input space, and then maps all correlations in these smaller 3D spaces, via regular (3x3 or 5x5) convolutions. In Inception-ResNet, Szegedy et al. combined the power of residual learning and inception block (He et al. 2015a; Szegedy et al. 2016a). In doing so, filter concatenation was replaced by the residual connection. Moreover, Szegedy et al. experimentally showed that Inception-V4 with residual connections (Inception-ResNet) has the same generalization power as plain Inception-V4 but with increased depth and width. However, they observed that Inception-ResNet converges more quickly than Inception-V4, which depicts that training with residual connections accelerates the training of Inception networks significantly.

## 4.3    Multi-Path based CNNs

Training of deep networks is a challenging task, and this has been the subject of recent research on deep networks. Deep CNNs generally perform well on complex tasks. However, they may suffer from performance degradation, gradient vanishing, or explosion problems, which are not caused by overfitting but instead by an increase in the depth (Hochreiter 1998; Dong et al. 2016). Vanishing gradient problem not only results in higher test error but also higher training error (Pascanu et al. 2012; Dong et al. 2016; Dauphin et al. 2017). For training deep networks, the concept of multi-path or cross-layer connectivity was proposed (Srivastava et al. 2015a; Larsson et al. 2016; Huang et al. 2017; Kuen et al. 2018). Multiple paths or shortcut connections can systematically connect one layer to another by skipping some intermediate layers to allow the specialized flow of information across the layers (Mao et al. 2016; Tong et al. 2017). Cross-layer connectivity partitions the network into several blocks. These paths also try to solve the vanishing gradient problem by making gradient accessible to lower layers. For this purpose, different types of shortcut connections are used, such as zero-padded, projection-based, dropout, skip connections, and 1x1 connections, etc.

### 4.3.1   Highway Networks

The increase in depth of a network improves performance mostly for complex problems, but it also makes training of the network difficult. In deep networks, due to a large number of layers, the backpropagation of error may result in small gradient values at lower layers. To solve this problem, in 2015, a new CNN architecture named Highway Networks was proposed based on the idea of cross-layer connectivity (Srivastava et al. 2015a). In Highway Networks, the unimpeded flow of information across layers is enabled by imparting two gating units within a layer (equation (5)). The idea of a gating mechanism was inspired by Long Short Term Memory (LSTM) based on Recurrent Neural Networks (RNN) (Mikolov et al. 2010; Sundermeyer et al.

2012). The aggregation of information by combining the $l^{th}$ layer and previous $l - j$ layers information creates a regularizing effect, making gradient-based training of very deep networks easy. This cross-layer connectivity enables the training of a network with more than 100 layers, even as deep as 900 layers with a stochastic gradient descent algorithm. Cross-layer connectivity for Highway Network is defined in equation (5 & 6).

$$\mathbf{F}_{l+1}^k = g_c(\mathbf{F}_l^k, \mathbf{k}_l) . g_{t_g}(\mathbf{F}_l^k, {}^{t_g}\mathbf{k}_l) . g_{c_g}(\mathbf{F}_l^k, {}^{c_g}\mathbf{k}_l) \tag{5}$$

$$g_{c_g}(\mathbf{F}_l^k, {}^{c_g}\mathbf{k}_l) = 1 - g_{t_g}(\mathbf{F}_l^k, {}^{t_g}\mathbf{k}_l) \tag{6}$$

In equation (5), $g_c(\mathbf{F}_l^k, \mathbf{k}_l)$ represents the working of the $l^{th}$ hidden layer, whereas $t_g$ and $c_g$ are two gates that decide the flow of information across the layers. When $t_g$ gate is open, $t_g = 1$ then transformed input is assigned to the next layer. Whereas, when the value of $t_g = 0$ then $c_g$ gate establishes an effect of information highway and input $\mathbf{F}_l^k$ of $l^{th}$ layer is directly assigned to the next layer $l+1$ without any transformation.

### 4.3.2 ResNet

In order to address the problems faced during training of deep networks, ResNet exploited the idea of bypass pathways used in Highway Networks (He et al. 2015a). Mathematical formulation of ResNet is expressed in equations (7, 8 & 9).

$$\mathbf{F}_{m+1}^{k'} = g_c(\mathbf{F}_{l \to m}^k, \mathbf{k}_{l \to m}) + \mathbf{F}_l^k \quad m \geq l \tag{7}$$

$$\mathbf{F}_{m+1}^k = g_a(\mathbf{F}_{m+1}^{k'}) \tag{8}$$

$$g_c(\mathbf{F}_{l \to m}^k, \mathbf{k}_{l \to m}) = \mathbf{F}_{m+1}^{k'} - \mathbf{F}_l^k \tag{9}$$

where, $g_c(\mathbf{F}_{l \to m}^k, \mathbf{k}_{l \to m})$ is a transformed signal, and $\mathbf{F}_l^k$ is an input of $l^{th}$ layer. In equation (7), $\mathbf{k}_{l \to m}$ shows the $k^{th}$ processing unit (kernel), whereas $l \to m$ suggests that the residual block can be consists of one or more than one hidden layers. Original input $\mathbf{F}_l^k$ is added to transformed signal ($g_c(\mathbf{F}_{l \to m}^k, \mathbf{k}_{l \to m})$) through bypass pathway (equation (7)) and thus results in an aggregated output $\mathbf{F}_{m+1}^{k'}$, which is assigned to the next layer after applying activation function $g_a(.)$.

Whereas, $(\mathbf{F}_{m+1}^{k^{i}} - \mathbf{F}_{l}^{k})$, returns a residual information, which is used to perform reference based optimization of weights. The distinct feature of ResNet is reference based residual learning framework. ResNet suggested that residual functions are easy to optimize and can gain accuracy for considerably increased depth.

ResNet introduced shortcut connections within layers to enable cross-layer connectivity; however, these connections are data-independent and parameter-free in comparison to the gates of Highway Networks. In Highway Networks, when a gated shortcut is closed, the layers represent non-residual functions. However, in ResNet, residual information is always passed, and identity shortcuts are never closed. Residual links (shortcut connections) speed up the convergence of deep networks, thus giving ResNet the ability to avoid gradient diminishing problems.

### 4.3.3 DenseNet

Similar to Highway Networks and ResNet, DenseNet was proposed to solve the vanishing gradient problem (Srivastava et al. 2015a; He et al. 2015a; Huang et al. 2017). The problem with ResNet was that it explicitly preserves information through additive identity transformations due to which many layers may contribute very little or no information. To address this problem, DenseNet used cross-layer connectivity but, in a modified fashion. DenseNet connected each preceding layer to the next coming layer in a feed-forward fashion; thus, feature-maps of all previous layers were used as inputs into all subsequent layers as expressed in equation (10 & 11).

$$\mathbf{F}_2^k = g_c(I_C, \mathbf{k}_1) \tag{10}$$

$$\mathbf{F}_l^k = g_k(\mathbf{F}_1^k, ..., \mathbf{F}_{l-1}^k) \tag{11}$$

where $\mathbf{F}_2^k$ and $\mathbf{F}_l^k$ are the resultant feature-maps of 1st and $l\text{-}1$th transformation layers and $g_k(.)$ is a function, which enables cross-layer connectivity by concatenating proceeding layers information before assigning to new transformation layer $l$. This establishes $\dfrac{l(l+1)}{2}$ direct connections in DenseNet, as compared to $l$ connections between a layer and its preceding layer in the traditional CNNs. It imprints the effect of cross-layer depthwise convolutions. As DenseNet concatenates the features of the previous layer instead of adding them, thus, the network may gain the ability to explicitly differentiate between information that is added to the network and information that is preserved. DenseNet has a narrow layer structure; however, it becomes parametrically expensive with an increase in a number of feature-maps. Information flow in the network improves by providing each layer direct access to the gradients through the loss function. Direct admittance to gradient incorporates a regularizing effect, which reduces overfitting on tasks with smaller training sets.

## 4.4    Width based Multi-Connection CNNs

During 2012-2015, the focus was mainly on exploiting the power of depth, along with the effectiveness of multi-pass regulatory connections in network regularization (Srivastava et al. 2015a; He et al. 2015a). However, Kawaguchi et al. reported that the width of the network is also important (Kawaguchi et al. 2019). Multilayer perceptron gained the advantage of mapping complex functions over perceptron by making parallel use of multiple processing units within a layer. This suggests that width is an essential parameter in defining principles of learning along with depth. Lu et al. (2017), and Hanin and Sellke (2017) have recently shown that NNs with ReLU activation function have to be wide enough to hold universal approximation property along with an increase in depth (Hanin and Sellke 2017). Moreover, a class of continuous functions on a compact set cannot be arbitrarily well approximated by an arbitrarily deep network, if the maximum width of the network is not larger than the input dimension (Lu et al. 2017b; Nguyen et al. 2018). Although, stacking of multiple layers (increasing depth) may learn diverse feature representations, but may not necessarily increase the learning power of the NN. One major problem linked with deep architectures is that some layers or processing units may not learn useful features. To tackle this problem, the focus of research shifted from deep and narrow architecture towards thin and wide architectures.

### 4.4.1   Wide ResNet

It is concerned that the main drawback associated with deep residual networks is the feature reuse problem in which some feature transformations or blocks may contribute very little to learning (Srivastava et al. 2015b). This problem was addressed by Wide ResNet (Zagoruyko and Komodakis 2016). Zagoruyko and Komodakis suggested that the main learning potential of deep residual networks is due to the residual units, whereas depth has a supplementary effect. Wide ResNet exploited the power of the residual blocks by making ResNet wide rather than deep (He et al. 2015a). Wide ResNet increased the width by introducing an additional factor *k*, which controls the width of the network. Wide ResNet showed that the widening of the layers might provide a more effective way of a performance improvement than by making the residual networks deep.

Deep networks improved representational capacity, but they have some demerits such as time-intensive training, feature reuse, and gradient vanishing and exploding problem. He et al. addressed feature reuse problem by incorporating dropout in residual blocks to regularize network effectively (He et al. 2015a). Similarly, Huang et al. introduced the concept of stochastic depth by exploiting dropouts to solve vanishing gradient and slow learning problems (Huang et al. 2016a). It was observed that even fraction improvement in performance might require the addition of many new layers. However, Zagoruyko and Komodakis (2016), empirically showed that though Wide ResNet was twice in a number of parameters as compared to ResNet, but can be trained in a better way than the deep networks (Zagoruyko and Komodakis 2016). Wide ResNet was based on the observation that almost all architectures before residual networks, including the most successful Inception and VGG, were wide as compared to ResNet. In Wide ResNet, learning is made effective by adding a dropout in between the convolutional layers rather than inside a residual block.

### 4.4.2 Pyramidal Net

In earlier deep CNN architectures such as AlexNet, VGG, and ResNet, due to the deep stacking of multiple convolutional layers, depth of feature-maps increases in subsequent layers. However, the spatial dimension decreases, as each convolutional layer or block is followed by a sub-sampling layer (Krizhevsky et al. 2012; Simonyan and Zisserman 2015; He et al. 2015a). Therefore, Han et al. argued that in deep CNNs, a drastic increase in the feature-map depth and, at the same time, the loss of spatial information limits the learning ability of CNN (Han et al. 2017). ResNet has shown remarkable results for image classification problems. However, in ResNet, the deletion of a residual block, where the dimension of both spatial and feature-map (channel) varies (feature-map depth increases, while spatial dimension decreases), generally deteriorates performance. In this regard, stochastic ResNet improved the performance by reducing information loss associated with the dropping of the residual unit (Huang et al. 2016a). To increase the learning ability of ResNet, Han et al. proposed the Pyramidal Net (Han et al. 2017). In contrast to drastic decrease in spatial width with an increase in depth by ResNet,

Pyramidal Net increases the width gradually per residual unit. This strategy enables pyramidal Net to cover all possible locations instead of maintaining the same spatial dimension within each residual block until down-sampling occurs. Because of a gradual increase in the depth of features map in a top-down fashion, it was named as pyramidal Net. In pyramidal network, depth of feature-maps is regulated by factor $l$, and is computed using equation (12).

$$d_l = \begin{cases} 16 & if \ l = 1, \\ \left\lfloor d_{l-1} + \frac{\lambda}{n} \right\rfloor & if \ 2 \leq l \leq n+1 \end{cases} \qquad (12)$$

where, $d_l$ denotes the dimensions of $l^{th}$ residual block and $n$ describes the number of the residual block, whereas $\lambda$ is a step size and $\frac{\lambda}{n}$ regulates the increase in depth. The depth regulating factor tries to distribute the burden of increase in depth of feature-maps. Residual connections were inserted in between the layers by using zero-padded identity mapping. The advantage of zero-padded identity mapping is that it needs less number of parameters as compared to the projection-based shortcut connection, hence may result in better generalization (Wang et al. 2019). Pyramidal Net uses two different approaches for the widening of the network, including addition and multiplication based widening. The difference between the two types of widening is that additive pyramidal structure increases linearly, whereas multiplicative one increases geometrically (Ioffe and Szegedy 2015; Xu et al. 2015a). However, a major problem with Pyramidal Net is that with the increase in width, a quadratic times increase in both space and time occurs.

### 4.4.3  Xception

Xception can be considered as an extreme Inception architecture, which exploits the idea of depthwise separable convolution (Chollet 2017). Xception modified the original inception block by making it wider and replacing the different spatial dimensions (1x1, 5x5, 3x3) with a single dimension (3x3) followed by a 1x1 convolution to regulate computational complexity.

The Architecture of the Xception block is shown in Fig. 8. Xception makes the network computationally efficient by decoupling spatial and feature-map (channel) correlation, which is mathematically expressed in equation (13 & 14). It works by first mapping the convolved output to low dimensional embeddings using 1x1 convolutions. It then spatially transforms it *n* times, where *n* is a width defining cardinality, which determines the number of transformations.

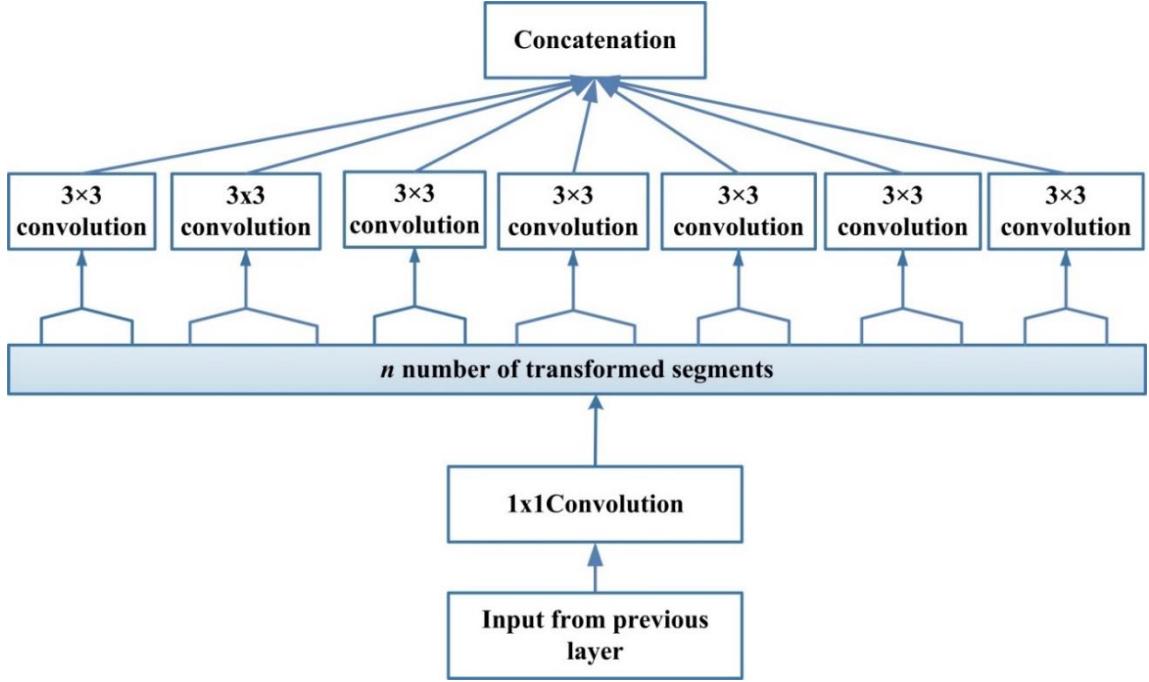

**Fig. 8** Xception building block and its *n* sets of transformation.

$$f_{l+1}^k(p,q) = \sum_{x,y} f_l^k(x,y).e_l^k(u,v) \tag{13}$$

$$\mathbf{F}_{l+2}^k = g_c(\mathbf{F}_{l+1}^k, \mathbf{k}_{l+1}) \tag{14}$$

In equation (14), $\mathbf{k}_l$ is a $k^{\text{th}}$ kernel of $l^{\text{th}}$ layer having depth one, which is spatially convolved across $k^{\text{th}}$ feature-map $\mathbf{F}_l^k$, where $(x, y)$ and $(u, v)$ show the spatial indices of feature-map and kernel respectively. In depthwise separable convolution, it is to be noted that number of kernels $K$ is equal to number of input feature-maps contrary to conventional convolutional layer where number of kernels are independent of previous layer feature-maps. Whereas $\mathbf{k}_{l+1}$ is $k^{\text{th}}$ kernel of (1x1) spatial dimension for $l+1^{\text{th}}$ layer, which performs depthwise convolution across output feature-maps $[\mathbf{F}_{l+1}^1, ..., \mathbf{F}_{l+1}^k, ..., \mathbf{F}_{l+1}^K]$ of $l^{\text{th}}$ layer, used as input of $l+1^{\text{th}}$ layer.

Xception makes computation easy by separately convolving each feature-map across spatial axes, which is followed by pointwise convolution (1x1 convolutions) to perform cross-channel correlation. In conventional CNN architectures; convolutional operation uses only one transformation segment, inception block uses three transformation segments, whereas in Xception number of transformation segments is equal to the number of feature-maps. Although the transformation strategy adopted by Xception does not reduce the number of parameters, it makes learning more efficient and results in improved performance.

### 4.4.4   ResNeXt

ResNeXt, also known as Aggregated Residual Transform Network, is an improvement over the Inception Network (Xie et al. 2017). Xie et al. exploited the concept of the split, transform, and merge in a powerful but simple way by introducing a new term; cardinality (Szegedy et al. 2015). Cardinality is an additional dimension, which refers to the size of the set of transformations (Han et al. 2018; Sharma and Muttoo 2018). The Inception network has not only improved the learning capability of conventional CNNs, but it also makes a network resource-efficient. However, due to the use of diverse spatial embedding's (such as the use of 3x3, 5x5, and 1x1 filter) in the transformation branch, each layer needs to be customized separately. ResNeXt utilized the deep homogenous topology of VGG and simplified GoogleNet architecture by fixing spatial resolution to 3x3 filters within the split, transform, and merge block. Whereas, it used residual learning to improve the convergence of deep and wide network (Simonyan and Zisserman 2015; Szegedy et al. 2015; He et al. 2015a). The building block for ResNeXt is shown in Fig. 9. ResNeXt used multiple transformations within a split, transform and merge block and defined these transformations in terms of cardinality. Xie et al. (2017) showed that an increase in cardinality significantly improves performance. The complexity of ResNeXt was regulated by applying low embedding's (1x1 filters) before 3x3 convolution, whereas training was optimized by using skip connections (Larsson et al. 2016).

### 4.4.5   Inception Family

Inception family of CNNs also comes under the class of width based methods (Szegedy et al. 2015, 2016a, b). In Inception networks, within a layer, varying sizes of the filters were used, which increased the output of the intermediate layers. The use of the different sizes of filters helps capture the diversity in high-level features. Salient characteristics of the Inception family are discussed in section 4.1.5 and 4.2.3.

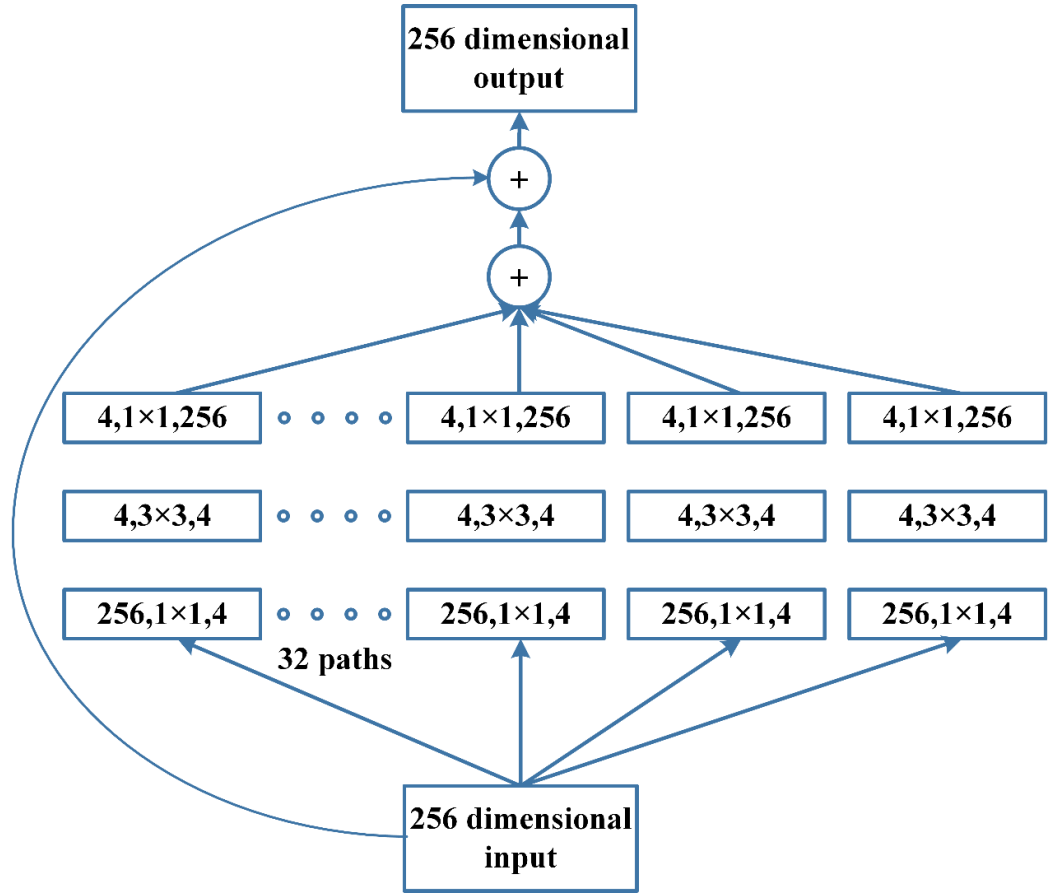

**Fig. 9** ResNeXt building block showing the different paths of transformation.

## 4.5 Feature-Map (Channel*FMap*) Exploitation based CNNs

CNN became popular for MV tasks because of its hierarchical learning and automatic feature extraction ability (LeCun et al. 2010). Feature selection plays a vital role in determining the performance of classification, segmentation, and detection modules. In CNN, features are dynamically selected by tuning the weights associated with a kernel also known as mask. Also, multiple stages of feature extraction are used, which can extract diverse types of features (known as feature-maps or channels in CNN). However, some of the feature-maps impart little or no role in object discrimination (Hu et al. 2018a). Enormous feature sets may create an effect of noise and thus lead to over-fitting of the network. This suggests that apart from network engineering, selection of feature-maps can play an important role in improving the generalization of the network. In this section, feature-maps and channels will be interchangeably used as many researchers have used the word channels for the feature-maps.

### 4.5.1 Squeeze and Excitation Network

Squeeze and Excitation Network (SE-Network) was reported by Hu et al. (Hu et al. 2018a). They proposed a new block for the selection of feature-maps (commonly known as channels) relevant to object discrimination. This new block was named as SE-block (shown in Fig. 10), which suppresses the less important feature-maps, but gives high weightage to the class specifying feature-maps. SE-Network reported a record decrease in error on the ImageNet dataset. SE-block is a processing unit that is designed generically and therefore, can be added in any CNN architecture before the convolution layer. The working of this block consists of two operations; squeeze and excitation. Convolution kernel captures information locally, but it ignores the contextual relation of features (correlation) that are outside of this receptive field (LeCun et al. 2015). Squeeze operation is performed to get a global view of feature-maps. The squeeze block generates feature-map wise statistics (also known as feature-map motifs or descriptors) by suppressing spatial information of the convolved input. As global average pooling has the potential to learn the extent of target object effectively (Lin et al. 2013; Zhou et al. 2016), therefore, it is employed by the squeeze operation $g_{sq}(.)$ using the following equation (15):

$$s_l^k = g_{sq}(\mathbf{F}_l^k) = \frac{1}{P \times Q} \sum_{p,q} f_l^k(p,q) \qquad (15)$$

where, $s_l^k$ represents a feature descriptor for $k^{\text{th}}$ feature-map of $l^{\text{th}}$ layer, and $P \times Q$ defines the spatial dimension of feature-map $\mathbf{F}_l^k$. Whereas output of squeeze operation $\mathbf{S}_l^K = \left[ s_l^1, ..., s_l^K \right]$ for $K$ number of convolved feature-maps for $l^{\text{th}}$ layer is assigned to the excitation operation $g_{ex}(.)$, which models motif-wise interdependencies by exploiting gating mechanism. Excitation operation assigns weights to feature-maps using two layer feed forward NN, which is mathematically expressed in equation (16).

$$y_{l+1}^k = g_{ex}(\mathbf{S}_l^k) = g_{s_g}(\mathbf{w}_2, g_t(\mathbf{S}_l^k, \mathbf{w}_1)) \qquad (16)$$

In equation (16), $y_{l+1}^k$ denotes weightage for input feature-map $\mathbf{F}_{l+1}^k$ of next layer ($l+1$), where $g_t(.)$ and $g_{s_g}(.)$ apply the ReLU based non-linear transformation and sigmoid gate, respectively. Similarly, $\mathbf{Y}_{l+1}^{K} = \left[ y_{l+1}^{1},...,y_{l+1}^{K} \right]$ shows the weightage for $K$ number of convolved feature-maps that are used to rescale them before assigning to the $l+I^{\text{th}}$ layer. In excitation operation, $\mathbf{w}_1$ and $\mathbf{w}_2$ both are used as transformation weight vectors and regulating factors to limit the model complexity and aid the generalization (LeCun 2007; Xu et al. 2015a). The output of the first hidden transformation in NN is preceded by the ReLU activation function, which inculcates non-linearity in motif responses. The gating mechanism is exploited in SE-block using the sigmoid activation function, which models the non-linear responses of the feature-maps and assigns a weight based on feature-map relevance (Zheng et al. 2017). SE-block adaptively recalibrates the feature-maps of each layer by multiplying convolved input with the motif responses.

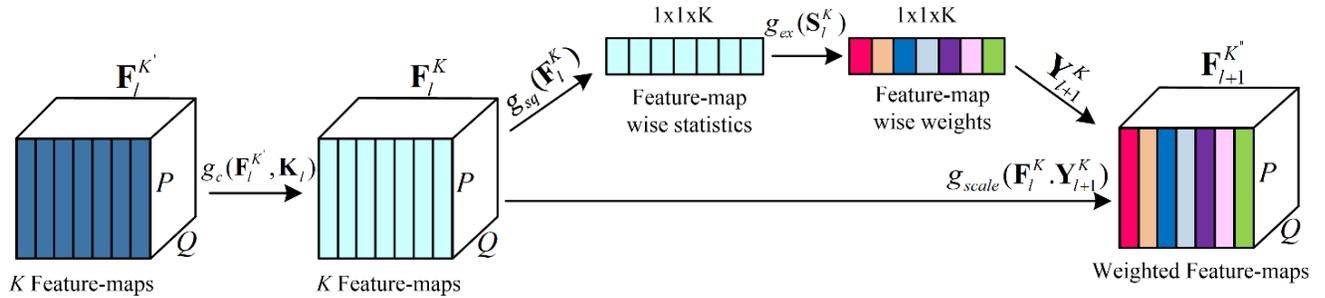

**Fig. 10** Squeeze and Excitation block showing the computation of masks for the recalibration of feature-maps that are commonly known as channels in literature.

### 4.5.2   Competitive Squeeze and Excitation Networks

Competitive Inner-Imaging Squeeze and Excitation for Residual Network also known as CMPE-SE Network was proposed by Hu et al. in 2018 (Hu et al. 2018b). Hu et al. used the idea of SE-block to improve the learning of deep residual networks (Hu et al. 2018a). SE-Network recalibrates the feature-maps based upon their contribution in class discrimination. However, the main concern with SE-Net is that in ResNet, it only considers the residual information for determining the weight of each feature-map (Hu et al. 2018a). This minimizes the impact of SE-block and makes ResNet information redundant. Hu et al. addressed this problem by generating feature-map wise motifs (statistics) from both residual and identity mapping based feature-maps. In this regard, global representation of feature-maps is generated using global average pooling operation (equation (17)), whereas relevance of feature-maps is estimated by establishing competition between feature descriptors of residual and identity mappings. This phenomena is termed as inner imaging (Hu et al. 2018b). CMPE-SE block not only models the relationship between residual feature-maps but also maps their relation with identity feature-map. The mathematical expression for CMPE-SE block is represented using the following equation:

$$\mathbf{S}_l^k, \mathbf{S}_{m+1}^k = g_{sq}(\mathbf{F}_l^k), g_{sq}(\mathbf{F}_{m+1}^{k^{'}}) \tag{17}$$

$$\mathbf{Y}_{m+1}^k = g_{ex}(g_k(\mathbf{S}_l^k, \mathbf{S}_{m+1}^k)) \tag{18}$$

$$\mathbf{F}_{m+1}^k = \mathbf{Y}_{m+1}^k . \mathbf{F}_{m+1}^{k^{'}} \tag{19}$$

where $\mathbf{F}_l^K$ and $\mathbf{F}_{m+1}^{K^{'}}$ are the identity and residual mapping of input $\mathbf{F}_l^K$ respectively. SE block is implemented by applying squeeze operation $g_{sq}(.)$ both on residual and the identity feature-maps and their receptive output is used as joint input of excitation operation $g_{ex}(.)$. Whereas $g_k(.)$ represents the concatenation operation. The output masks of excitation operation (equation (18)) are multiplied with residual information (equation (19)) to rebuild each feature-map importance. The backpropagation algorithm thus tries to optimize the competition between identity and residual feature-maps and the relationship between all feature-maps in the residual block.

## 4.6    Channel$_{(Input)}$ Exploitation based CNNs

Image representation plays an important role in determining the performance of the image processing algorithms, including both conventional and deep learning algorithms. A good representation of the image is one that can define the salient features of an image from a compact code. In MV tasks, various types of conventional filters are applied to extract different levels of information for a single type of image (Lowe 2004; Dollár et al. 2009). These diverse representations are then used as an input of the model to improve performance (Do and Vetterli 2005; Oquab et al. 2014). Now CNN is a compelling feature learner that can automatically extract discriminating features depending upon the problem (Yang et al. 2019). However, the learning of CNN relies on input representation. The lack of diversity and the absence of class discernable information in the input may affect CNN's performance as a discriminator. For this purpose, the concept of channel boosting (input channel dimension) using auxiliary learners is introduced in CNNs to boost the representation of the network (Khan et al. 2018a).

### 4.6.1 Channel Boosted CNN using TL

In 2018, Khan et al. proposed a new CNN architecture named as Channel boosted CNN (CB-CNN)based on the idea of boosting the number of input channels for improving the representational capacity of the network (Khan et al. 2018a). The Block diagram of CB-CNN is shown in Fig. 11. Channel boosting is performed by artificially creating extra channels (known as auxiliary channels) through auxiliary deep generative models and then exploiting it through the deep discriminative models. CB-CNN is mathematically expressed in equation (20 & 21).

$$\mathbf{I}_B = g_k\left(\mathbf{I}_C, \left[\mathbf{A}_1, ..., \mathbf{A}_M\right]\right) \tag{20}$$

$$\mathbf{F}_l^k = g_c(\mathbf{I}_B, \mathbf{k}_l) \tag{21}$$

In equation (20), $\mathbf{I}_C$ represents the original input channels, where $\mathbf{A}_M$ is an artificial channel generated by $M^{\text{th}}$ auxiliary learner. Whereas $g_k(.)$ is used as a combiner function that concatenates the original input channels with auxiliary channels to generates the channel boosted input $\mathbf{I}_B$ for the discriminator. Equation (21) shows the $k^{\text{th}}$ resultant feature-map $\mathbf{F}_l^k$, which is generated by convolving the boosted input $\mathbf{I}_B$ with kernel $\mathbf{k}_l$ of $l^{\text{th}}$ layer.

Bengio et al. in 2013, empirically showed that data representation plays an important role in determining the performance of a classifier, as different representations may present different aspects of information (Bengio et al. 2013). For improving the representation of the data, Khan et al. exploited the power of TL and deep generative learners (Qiang Yang et al. 2008; Vincent et al. 2008; Hamel and Eck 2010). Generative learners attempt to characterize the data generating distribution during the learning phase. In CB-CNN, AEs are used as the generative learners to learn explanatory factors of variation behind the data. The concept of inductive TL is used in a novel way to build a boosted input representation by augmenting learned distribution of the input data with the original channel space (input channels). CB-CNN encodes channel-boosting phase into a generic block, which is inserted at the start of a deep network. CB-CNN provides the concept that TL can be used at both generation and discrimination stages. The significance of the study is that multi-deep learners are used, where generative learning models are used as auxiliary learners. These leaners enhance the representational capacity of deep CNN based discriminator.

Although the potential of the channel boosting was only evaluated by inserting a boosting block at the start, however, Khan et al. suggested that this idea can be extended by providing auxiliary channels at any layer in the deep architecture. CB-CNN has also been evaluated on the medical image dataset (Aziz et al. 2020), where it has shown improved results, as shown in Table 4.

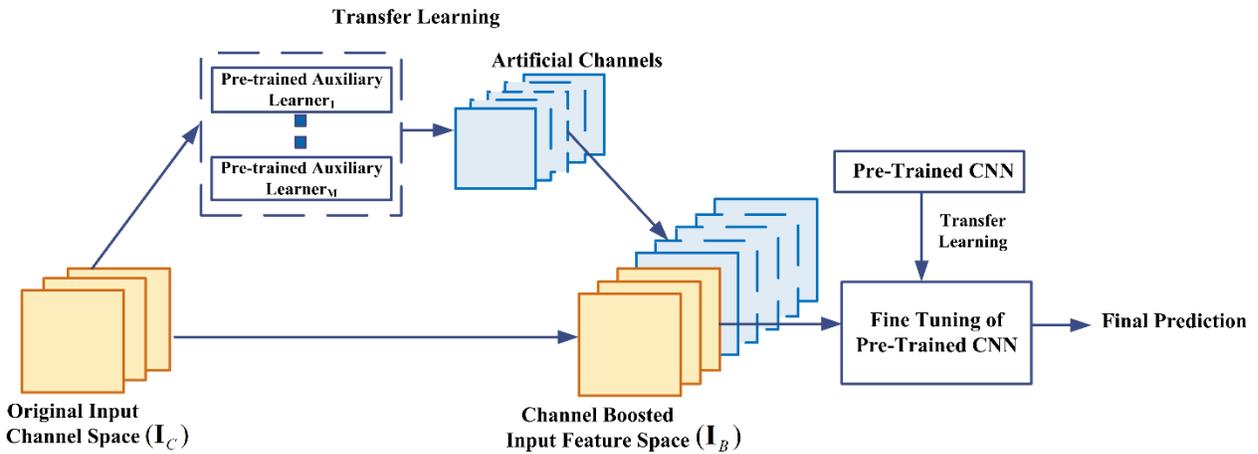

**Fig. 11** Basic architecture of CB-CNN showing the deep auxiliary learners for creating artificial channels.

**Table 4** Results of CNN and CB-CNN on mitosis dataset.

| CNN Architecture | F-score |
|---|---|
| 26 layers deep CNN | 0.47 |
| 26 layers deep CB-CNN | 0.53 |
| VGG | 0.55 |
| CB-VGG | 0.71 |
| ResNet | 0.44 |
| CB-ResNet | 0.54 |

## 4.7 Attention based CNNs

Different levels of abstractions have an important role in defining the discrimination power of the NN. In addition to learning of multiple hierarchies of abstractions, focusing on features relevant to the context also plays a significant role in image localization and recognition. In the human visual system, this phenomenon is referred to as attention. Humans view the scene in a succession of partial glimpses and pay attention to context-relevant parts. This process not only serves to focus selected regions but also deduces different interpretations of objects at that location and thus helps in capturing visual structure in a better way. A more or less similar kind of interpretability is added into RNN and LSTM (Mikolov et al. 2010; Sundermeyer et al. 2012). RNN and LSTM networks exploit attention modules for the generation of sequential data, and the new samples are weighted based on their occurrence in previous iterations. The concept of attention was incorporated into CNN, by various researchers to improve representation and overcome the computational limits. This idea of attention also helps in making CNN intelligent enough to recognize objects even from cluttered backgrounds and complex scenes.

### 4.7.1 Residual Attention Neural Network

Wang et al. proposed a Residual Attention Network (RAN) to improve the feature representation of the network (Wang et al. 2017a). The motivation behind the incorporation of attention in CNN was to make a network capable of learning object aware features. RAN is a feed-forward CNN, which was built by stacking residual blocks with attention module. The attention module is branched off into trunk and mask branches that adopt bottom-up, top-down learning strategy. The assembly of two different learning strategies into the attention module enables fast feed-forward processing and top-down attention feedback in a single feed-forward process. The bottom-up feed-forward structure produces low-resolution feature-maps with strong semantic information. Whereas, top-down architecture produces dense features to make an inference of each pixel.

In the previously proposed studies, a top-down, bottom-up learning strategy was used by Restricted Boltzmann Machines (Salakhutdinov and Larochelle 2010). Similarly, Goh et al. exploited the top-down attention mechanism as a regularizing factor in Deep Boltzmann Machine during the reconstruction phase of the training. The top-down learning strategy globally optimizes the network in such a way that it gradually output the maps to input during the learning process (Hinton et al. 2006; Salakhutdinov and Larochelle 2010; Goh et al. 2013). The attention module in RAN generates object aware soft mask $g_{sm}(.)$ at each layer for input feature-map $\mathbf{F}_l^K$ (Xu et al. 2015b). Soft mask $g_{sm}(.)$ assigns attention towards object using equation (22) by recalibrating trunk branch $g_{tm}(\mathbf{F}_l^K)$ output and thus, behaves like a control gate for every neuron output.

$$g_{am}(\mathbf{F}_l^K) = g_{sm}(\mathbf{F}_l^K).g_{tm}(\mathbf{F}_l^K) \tag{22}$$

In one of the previous studies, Transformation network (Jaderberg et al. 2015; Li et al. 2018) also exploited the idea of attention in a simple way by incorporating it with convolution block, but the main problem was that attention modules in Transformation network are fixed and cannot adapt to changing circumstances. RAN was made efficient towards recognition of cluttered, complex, and noisy images by stacking multiple attention modules. Hierarchical organization of RAN endowed the ability to adaptively assign weight to each feature-map based on their relevance in the layers (Wang et al. 2017a). Learning of deep hierarchical structure was supported through residual units. Moreover, three different levels of attention: mixed, channel, and spatial attention were incorporated, thus leveraging the capability to capture object-aware features at different levels (Wang et al. 2017a).

### 4.7.2   Convolutional Block Attention Module

The significance of attention mechanism and feature-map exploitation is validated through RAN and SE-Network (Wang et al. 2017a; Hu et al. 2018a). In this regard, Woo et al. came up with new attention-based CNN, named as Convolutional Block Attention Module (CBAM) (Woo et al. 2018). CBAM is simple in design and similar to SE-Network. SE-Network only considers the contribution of feature-maps in image classification, but it ignores the spatial locality of the object in images. The spatial location of the object has a vital role in object detection.

CBAM infers attention maps sequentially by first applying feature-map (channel) attention and then spatial attention, to find the refined feature-maps. In literature, generally, 1x1 convolution and pooling operations are used for spatial attention. Woo et al. showed that the pooling of features along the spatial axis generates an efficient feature descriptor. CBAM concatenates average pooling operation with max-pooling, which generates a strong spatial attention map. Likewise, feature-map statistics were modeled using a combination of max-pooling and global average-pooling operation. Woo et al. showed that max-pooling could provide the clue about distinctive object features, whereas the use of global average pooling returns suboptimal inference of feature-map attention. The exploitation of both average-pooling and max-pooling improves the representational power of the network. These refined feature-maps not only focus on the important part but also increase the representational power of the selected feature-maps. Woo et al. empirically showed that the formulation of a 3D attention map via the serial learning process helps in the reduction of the parameters as well as computational cost. Due to the simplicity of CBAM, it can be integrated easily with any CNN architecture.

### 4.7.3 Concurrent Spatial and Channel Excitation Mechanism

In 2018, Roy et al. extended the work of Hu et al. by incorporating the effect of spatial information in combination with feature-map (channel) information to make it applicable to segmentation tasks (Roy et al. 2018; Hu et al. 2018a). They introduced three different modules: (i) squeezing spatially and exciting feature-map information (cSE), (ii) squeezing feature-map and exciting spatial information (sSE), and (iii) concurrent squeeze and excitation of spatial and feature-map information (scSE). In this work, AE based convolutional NN was used for segmentation, whereas proposed modules were inserted after the encoder and decoder layers. In the cSE module, the same concept as that of SE-block is exploited. In this module, the scaling factor is derived based on the combination of feature-maps used for object detection. As spatial information has an important role in segmentation, therefore in the sSE module, the spatial locality has been given more importance than feature-map information. For this purpose, different combinations of feature-maps are selected and exploited spatially to use them for segmentation. In the last module, scSE, attention to each feature-map, is assigned by deriving scaling factor both from spatial and feature-map information and thus to highlight the object-specific feature-maps [117] selectively (Roy et al. 2018).

## 5    Applications of CNNs

CNNs have been successfully applied to different ML related tasks, namely object detection, recognition, classification, regression, segmentation, etc., (Batmaz et al. 2019; Chouhan and Khan 2019; Wahab et al. 2019). However, CNN generally needs a large amount of data for learning. All of the areas mentioned earlier in which CNN has shown tremendous success have relatively sufficient labeled data, such as traffic sign recognition, segmentation of medical images, and the detection of faces, text, pedestrians, and humans in natural images. Some of the interesting applications of CNN are discussed below.

### 5.1    CNN based computer vision and related applications

Computer vision (CV) focuses on developing an artificial system that can process visual data, including images and videos and can effectively understand, and extract useful information from it. CV encompasses a number of application areas such as face recognition, pose estimation, activity recognition, etc.

Face recognition is one of the difficult tasks in the CV. Face recognition systems have to cope with variations such as caused by illumination, change in pose, and different facial expressions. Farfade et al. (2015) proposed deep CNN for detecting faces from different poses as well as from occluded faces (Farfade et al. 2015). In another work, Zhang et al. performed face detection using a new type of multitasking cascaded CNN (Zhang et al. 2016). Zhang's technique showed good results when compared to state-of-the-art techniques (Li et al. 2015; Ranjan et al. 2015; Yang et al. 2015).

Human pose estimation is one of the challenging tasks related to CV because of the high variability in body pose. Li et al. (2014) proposed a heterogeneous deep CNN based pose estimation related technique (Li et al. 2014). In Li's technique, empirical results have shown that the hidden neurons can learn the localized part of the body. Similarly, another cascade based CNN technique is proposed by Bulat et al. (Bulat and Tzimiropoulos 2016). In their cascaded architecture, first heat maps are detected, whereas, in the second phase, regression is performed on the detected heat maps.

Action recognition is one of the important areas of activity recognition. The difficulties in developing an action recognition system are to solve the translations and distortions of features in different patterns, which belong to the same action class. Earlier approaches involved the construction of motion history images, the use of Hidden Markov Models, action sketch generation, etc. Recently, Wang et al. (Wang et al. 2017b) proposed a three dimensional CNN architecture in combination with LSTM for recognizing different actions from video frames. Experimental results have shown that Wang's technique outperforms other activity recognition based techniques (Wang and Schmid 2013; Simonyan and Zisserman 2014; Donahue et al. 2015; Sun et al. 2015; Tran et al. 2015). Similarly, another three dimensional CNN based action recognition system is proposed by Ji et al. (Ji et al. 2010). In Ji's work, three-dimensional CNN is used to extract features from multiple channels of input frames. The final action recognition based model is developed on combined extracted feature space. The proposed three dimensional CNN model is trained in a supervised way and can perform activity recognition in real-world applications.

## 5.2    CNN based natural language processing

Natural Language Processing (NLP) converts language into a presentation that can easily be exploited by any computer. Although RNNs are very suitable for NLP applications, however, CNNs have also been utilized in NLP based applications such as language modeling, and analysis, etc. Especially, language modeling or sentence molding has taken a twist after the introduction of CNN as a new representation learning algorithm. Sentence modeling is performed to know semantics of the sentences and thus offer new and appealing applications according to customer requirements. Traditional methods of information retrieval analyze data, based on words or features, but ignore the core of the sentence. Kalchbrenner et al. (2014) proposed a dynamic CNN and dynamic *k-max* pooling during training. This approach finds the relations between words without taking into account any external source like parser or vocabulary (Kalchbrenner et al. 2014). In a similar way, Collobert and Weston (2008) proposed CNN based architecture that can perform various MLP related tasks at the same time as chunking, language modeling, recognizing name-entity, and role modeling related to semantics (Collobert and Weston 2008). In another work, Hu et al. proposed a generic CNN based architecture that performs matching between two sentences and thus can be applied to different languages (Hu et al. 2011).

## 5.3    CNN based object detection and segmentation

Object detection focuses on identifying different objects in images. Recently, R-CNN has been widely used for object detection. Ren et al. (2015) proposed an improvement over R-CNN named as fast R-CNN for object detection (Ren et al. 2015). In their work, a fully connected convolutional neural network is used to extract feature space that can simultaneously detect the boundary and score of objects located at different positions. Similarly, Dai et al. (2016) proposed region-based object detection using fully connected CNN (Dai et al. 2016). In Dai's work, results are reported on the PASCAL VOC image dataset. Another object detection technique is reported by Gidaris et al. (Gidaris and Komodakis 2015), which is based on multi-region based deep CNN that helps to learn the semantic aware features. In Gidaris's approach, objects are detected with high accuracy on PASCAL VOC 2007 and 2012 dataset. Recently, AE based CNN architectures have shown success in segmentation tasks. In this regard, various interesting CNN architectures have been reported for both semantic and instance-based segmentation tasks such as FCN,

SegNet, Mask R-CNN, U-Net etc., (Ronneberger et al. 2015; Badrinarayanan et al. 2017; He et al. 2017; Zhang et al. 2018b).

## 5.4    CNN based image classification

CNN has been widely used for image classification (Levi and Hassner 2009; Long et al. 2012; Sermanet et al. 2012). One of the primary applications of CNN is in medical images, especially for the diagnosis of cancer using histopathological images (Cireşan et al. 2013). Recently, Spanhol et al. (2016) used CNN for the diagnosis of breast cancer images, and results are compared against a network trained on a dataset containing handcrafted descriptors (Spanhol et al. 2016a, b). Another recently proposed CNN based technique for breast cancer diagnosis is developed by Wahab et al. (Wahab et al. 2017). In Wahab's work, two phases are involved. In the first phase, hard non-mitosis examples are identified. In the second phase, data augmentation is performed to cope with the class skewness problem. Similarly, Ciresan et al. (Cireşan et al. 2012) used the German benchmark dataset related to a traffic sign signal. They designed CNN based architecture that performed traffic sign classification related task with a good recognition rate.

## 5.5    CNN based speech recognition

Deep CNN is mostly considered as the best option to deal with image processing applications, however; recent studies have shown that it also performs well on speech recognition tasks. Hamid et al. reported a CNN based speaker-independent speech recognition system (Abdel-Hamid et al. 2012). Experimental results showed a ten percent reduction in error rate in comparison to the earlier reported methods (Dahl et al. 2010; Mohamed et al. 2012). In another work, various CNN architectures, which are either based on the full or limited number of weight sharing within the convolution layer, are explored (Abdel-Hamid et al. 2013). Furthermore, the performance of CNN is also evaluated after the initialization of the network using the pre-training phase (Mohamed et al. 2012). Experimental results showed that almost all of the explored architectures yield good performance on phone and vocabulary recognition related tasks. Nowadays, the utilization of CNNs for speech emotion recognition is also gaining attention. Huang et al. used CNN in combination with LSTM for recognizing emotions of speech. In Huang's approach, CNN was trained both on verbal and nonverbal segments of speech, and CNN learned features were used by LSTM for recognizing speech emotions (Huang et al. 2019).

## 5.6 CNN based video processing

In video processing techniques, temporal and spatial information from videos is exploited. Many researchers have used CNN to solve various video processing-related problems (Tong et al. 2015; Frizzi et al. 2016; Shi et al. 2017; Ullah et al. 2017; Wang et al. 2017b). Tong et al. proposed CNN based short boundary detection system. In Tong's approach, TAGs are generated using CNN (Tong et al. 2015). During the experiment, the merging of TAGs against one-shot is performed to annotate video against that shot. Similarly, Wang et al. used 3-D CNN along with LSTM to recognize action within the video (Wang et al. 2017b). In another technique, Frizzi et al. used CNN for detecting smoke and fire within the video (Frizzi et al. 2016). In Frizzi's approach, CNN architecture not only extracts salient features but also performs the classification task. In the field of action recognition, the gathering of spatial and temporal information is considered as a tedious task. In order to overcome the deficiencies of traditional feature descriptors, Tian et al. (Shi et al. 2017) proposed a three stream-based structure, which is capable of extracting spatial-temporal features along with short and long term motion within the video. Similarly, in another technique, CNN, in combination with bi-directional LSTM, is used for recognizing action from the video (Ullah et al. 2017). Their approach comprises of two phases. In the first phase, features are extracted from the sixth frame of the videos. In the second phase, sequential information between features of the frame is exploited using the bi-directional LSTM framework.

## 5.7 CNN for low resolution images

In the field of ML, different researchers have used CNN based image enhancement techniques for enhancing the resolution of the images (Chevalier et al. 2015; Peng et al. 2016; Kawashima et al. 2017). Peng et al. used deep CNN based approach, which categorizes the objects in images having low resolution (Peng et al. 2016). Similarly, Chevalier et al. introduced LR-CNN for low-resolution image classification (Chevalier et al. 2015). Another, the deep learning based technique is reported by Kawashima et al., in which convolutional layers, along with a layer of LSTM is used to recognize action from thermal images of low resolution (Kawashima et al. 2017).

## 5.8    CNN for resource limited systems

Despite CNN's high computational cost, it has been successfully utilized in developing different ML based embedded systems (Bettoni et al. 2017; Lee et al. 2017; Xie et al. 2018). Lee et al. developed the number plate recognition system, which is capable of quickly recognizing the number on the license plate (Lee et al. 2017). In Lee's technique, the deep learning based embedded recognition system comprises of simple AlexNet architecture. In order to address power efficiency and portability for embedded platforms, Bettoni et al. implemented CNN on the FPGA platform (Bettoni et al. 2017). In another technique, the FPGA embedded platform is used for efficiently performing different CNN based ML tasks (Xie et al. 2018). Similarly, resource-limited CNN architectures such as MobileNet, ShuffleNet, ANTNets, etc. are highly applicable for mobile devices (Howard et al. 2017; Zhang et al. 2018a; Xiong et al. 2019). Shakeel et al. developed a real-time based driver drowsiness detection application for smart devices such as android phones. They used MobileNet architecture in combination with SSD to exploit the benefit of the lightweight architecture of MobileNet that can be easily deployed on resource-constrained hardware and can learn enriched representation from the incoming video (Shakeel et al. 2019).

## 5.9    CNN for 1D-Data

CNN has not only shown good performance on images but also on 1D-data. The use of 1D-CNN as compared to other ML methods is becoming popular because of its good feature extraction ability. Vinayakumar et al. used 1D-CNN in combination with RNN, LSTM, and gated recurrent units for intrusion detection in network traffic (Vinayakumar et al. 2017). They evaluated the performance of the proposed models on the KDDCup 99 dataset consisting of network traffic of TCP/IP packets and showed that CNN significantly surpasses the performance of classical ML models. Abdeljaber et al. showed that 1D-CNN could be used for real-time structural damage detection problem (Abdeljaber et al. 2017). They developed an end-to-end system that can automatically extract damage-sensitive features from accelerated signals for detection purposes. Similarly, Yildirim et al. showed the successful use of CNN for the 1D biomedical dataset.

Yildirim et al. developed 16 layers deep 1D-CNN based application for mobile devices and a cloud-based environment for detecting cardiac irregularity in ECG signals. They achieved 91.33% accuracy on the MIT-BIH Arrhythmia database (Yıldırım et al. 2018).

**Table 5a** Major challenges associated with implementation of Spatial exploitation based CNN architectures.

| Spatial Exploitation | As convolutional operation considers the neighborhood (correlation) of input pixels, therefore different levels of correlation can be explored by using different filter sizes. | |
|---|---|---|
| **Architecture** | **Strength** | **Gaps** |
| LeNet | • Exploited spatial correlation to reduce the computation and number of parameters<br>• Automatic learning of feature hierarchies | • Poor scaling to diverse classes of images<br>• Large size filters<br>• Low level feature extraction |
| AlexNet | • Low, mid and high-level feature extraction using large and small size filters on initial (5x5 and 11x11) and last layers (3x3)<br>• Give an idea of deep and wide CNN architecture<br>• Introduced regularization in CNN<br>• Started parallel use of GPUs as an accelerator to deal with complex architectures | • Inactive neurons in the first and second layers<br>• Aliasing artifacts in the learned feature-maps due to large filter size |
| ZfNet | • Introduced the idea of parameter tuning by visualizing the output of intermediate layers<br>• Reduced both the filter size and stride in the first two layers of AlexNet | • Extra information processing is required for visualization |
| VGG | • Proposed an idea of effective receptive field<br>• Gave the idea of simple and homogenous topology | • Use of computationally expensive fully connected layers |
| GoogLeNet | • Introduced the idea of using Mutiscale Filters within the layers<br>• Gave a new idea of split, transform, and merge<br>• Reduce the number of parameters by using bottleneck layer, global average-pooling at last layer and Sparse Connections<br>• Use of auxiliary classifiers to improve the convergence rate | • Tedious parameter customization due to heterogeneous topology<br>• May lose the useful information due to representational bottleneck |

**Table 5b** Major challenges associated with implementation of Depth based CNN architectures.

| Depth | With the increase in depth, the network can better approximate the target function with a number of nonlinear mappings and improved feature representations. Main challenge faced by deep architectures is the problem of vanishing gradient and negative learning. | |
|---|---|---|
| **Architecture** | **Strength** | **Gaps** |
| Inception-V3 | • Exploited asymmetric filters and bottleneck layer to lessen the computational cost of deep architectures | • Complex architecture design<br>• Lack of homogeneity |
| Highway Networks | • Introduced training mechanism for deep networks<br>• Used auxiliary connections in addition to direct connections | • Parametric gating mechanism, difficult to implement |
| Inception-ResNet | • Combined the power of residual learning and inception block | - |
| Inception-V4 | • Deep hierarchies of features, multilevel feature representation | • Slow in learning |
| ResNet | • Decreased the error rate for deeper networks<br>• Introduced the idea of residual learning<br>• Alleviates the effect of vanishing gradient problem | • A little complex architecture<br>• Degrades information of feature-map in feed forwarding<br>• Over adaption of hyper-parameters for specific task, due to the stacking of same modules |

**Table 5c** Major challenges associated with implementation of Multi-Path based CNN architectures.

| Multi-Path | Shortcut paths provides the option to skip some layers. Different types of the shortcut connections used in literature are zero padded, projection, dropout, 1x1 connections, etc. | |
|---|---|---|
| **Architecture** | **Strength** | **Gaps** |
| Highway Networks | • Mitigates the limitations of deep networks by introducing cross layer connectivity. | • Gates are data dependent and thus may become parameter expensive |
| ResNet | • Use of identity based skip connections to enable cross layer connectivity<br>• Information flow gates are data independent and parameter free<br>• Can easily pass the signal in both directions, forward and backward | • Many layers may contribute very little or no information<br>• Relearning of redundant feature-maps may happen |
| DenseNet | • Introduced depth or cross-layer dimension<br>• Ensures maximum data flow between the layers in the network<br>• Avoid relearning of redundant feature-maps<br>• Low and high level both features are accessible to decision layers | • Large increase in parameters due to increase in number of feature-maps at each layer |

**Table 5d** Major challenges associated with implementation of Width based CNN architectures.

| Width | Earlier, it was assumed that to improve accuracy, the number of layers have to be increased. However, by increasing the number of layers, the vanishing gradient problem arises and training might get slow. So, the concept of widening a layer was also investigated. | |
|---|---|---|
| **Architecture** | **Strength** | **Gaps** |
| Wide ResNet | • Showes the effectiveness of parallel use of transformations by increasing the width of ResNet and decreasing its depth<br>• Enables feature reuse<br>• Have shown that dropouts between the convolutional layer are more effective | • Over fitting may occur<br>• More parameters than thin deep networks |
| Pyramidal Net | • Introduces the idea of increasing the width gradually per unit<br>• Avoids rapid information loss<br>• Covers all possible locations instead of maintaining the same dimension till last unit | • High spatial and time complexity<br>• May become quite complex, if layers are substantially increased |
| Xception | • Introduce the concept that learning across 2D followed by 1 D is easier than to learn filters in 3 D space<br>• Depth-wise separable convolution is introduced<br>• Use of cardinality to learn good abstractions | • High computational cost |
| Inception | • Varying size filters inside inception module increases the output of the intermediate layers<br>• Varying size filters are helpful to capture the diversity in high-detail images | • Increase in space and time complexity |
| ResNeXt | • Introduced cardinality to avail diverse transformations at each layer<br>• Easy parameter customization due to homogenous topology<br>• Uses grouped convolution | • High computational cost |

**Table 5e** Major challenges associated with implementation of Feature-Map exploitation based CNN architectures.

| Feature-Map Selection | As the deep learning topology is extended, more and more features maps are generated at each step. Many of the Feature-maps might be important for classification task, others might redundant or less important. Hence, feature-map selection is another important dimension in deep learning architectures. | |
|---|---|---|
| **Architecture** | **Strength** | **Gaps** |
| Squeeze and Excitation Network | • It is a block-based concept<br>• Introduced a generic block that can be added easily in any CNN model due to its simplicity<br>• Squeezes less important features and vice versa | • In ResNet, it only considers the residual information for determining the weight of each channel |
| Competitive Squeeze and Excitation Networks | • Uses feature-map wise statistics from both residual and identity mapping based features<br>• Makes a competition between residual and identity feature-maps | • Doesn't support the concept of attention |

**Table 5f** Major challenges associated with implementation of Channel Boosting based CNN architectures.

| Channel Boosting | The learning of CNN also relies on the input representation. The lack of diversity and absence of class discernable information in the input may affect CNN performance. For this purpose, the concept of channel boosting (input channel dimension) using auxiliary learners is introduced in CNN to boost the representation of the network (Khan et al. 2018a). | |
|---|---|---|
| **Architecture** | **Strength** | **Gaps** |
| Channel Boosted CNN using Transfer Learning | • It boosts the number of input channels for improving the representational capacity of the network<br>• Inductive Transfer Learning is used in a novel way to build a boosted input representation for CNN | • Increases in computational load may happen due to the generation of auxiliary channels |

**Table 5g** Major challenges associated with implementation of Attention based CNN architectures.

| Attention | Attention Networks advantages to choose which patch is the area of the focus or most important in an image | |
|---|---|---|
| **Architecture** | **Strength** | **Gaps** |
| Residual Attention Neural Network | • Generates attention aware feature-maps<br>• Easy to scale up due to residual learning<br>• Provides different representations of the focused patches<br>• Adds soft weights on features using bottom up top-down feedforward attention | • Complex model |
| Convolutional Block Attention Module | • CBAM is a generic block designed for feed forward convolutional neural networks.<br>• Generate both feature-map and spatial attention in a sequential manner<br>• Channel attention maps help what to focus.<br>• Spatial attention helps where to focus.<br>• Increases efficient flow of information.<br>• Uses global average pooling and max pool simultaneously. | • Increase in computational load may happen |

# 6    CNN challenges

Deep CNNs have achieved good performance on data that either is of the time series nature or follows a grid-like topology. However, there are also some other challenges, where deep CNN architectures have been put to tasks. Major challenges associated with different CNN architectures are mentioned in Table 5a-g. The different researchers related to the performance of CNN on different ML tasks have interesting discussions. Some of the challenges faced during the training of deep CNN models are given below:

- Deep CNNs are generally like a black box and thus may lack in interpretation and explanation. Therefore, sometimes it is difficult to verify them.

- Szegedy et al. (2013) showed that training of CNN on noisy image data could cause an increase of misclassification error (Szegedy et al. 2014). The addition of the small quantity of random noise in the input image is capable of fooling the network in such a way that the model will classify the original and its slightly perturbed version differently.

- Each layer of CNN automatically tries to extract better and problem-specific features related to the task. However, for some tasks, it is imperative to know the nature of features extracted by the deep CNNs before classification. The idea of feature visualization in CNNs can help in this direction. Similarly, Hinton reported that lower layers should handover their knowledge only to the relevant neurons of the next layer. In this regard, Hinton proposed an interesting Capsule Network approach (de Vries et al. 2016; Hinton et al. 2018).

- Deep CNNs are based on supervised learning mechanisms, and therefore, the availability of large and annotated data is required for its proper learning. In contrast, humans can learn and generalize from a few examples.

- Hyper-parameter selection highly influences the performance of CNN. A little change in the hyper-parameter values can affect the overall performance of a CNN. That is why careful selection of hyper-parameters is a major design issue that needs to be addressed through some suitable optimization strategy.

- The efficient training of CNN demands powerful hardware resources such as GPUs. However, it is still needed to employ CNNs in embedded and smart devices efficiently. A few applications of deep learning in embedded systems are wound intensity correction, law enforcement in smart cities, etc., (Hinton et al. 2011, 2012a; Lu et al. 2017a).

- In vision-related tasks, one shortcoming of CNN is that it is generally unable to show good performance when used to estimate the pose, orientation, and location of an object. In 2012, AlexNet solved this problem to some extent by introducing the concept of data augmentation. Data augmentation can help CNN in learning diverse internal representations, which ultimately may lead to improved performance.

## 7    Future directions

The exploitation of different innovative ideas in CNN architectural design has changed the direction of research, especially in image processing and CV. Good performance of CNN on a grid-like topological data presents it as a powerful representational model for images. Architectural design of CNN is a promising research field and in future, it is likely to be one of the most widely used AI techniques.

- Ensemble learning is one of the prospective areas of research in CNNs (Marmanis et al. 2016; Ahmed et al. 2019). The combination of multiple and diverse architectures can aid the model in improving generalization and robustness on diverse categories of images by extracting different levels of semantic representations. Similarly, concepts such as batch normalization, dropout, and new activation functions are also worth mentioning.

- The potential of a CNN as a generative learner is exploited in image segmentation tasks, where it has shown good results (Kahng et al. 2019). The exploitation of generative learning capabilities of CNNs at feature extraction stages can boost the representational power of the model. Similarly, new paradigms are needed that can enhance the learning capacity of CNN by incorporating informative feature-maps that can be learned using auxiliary learners at the intermediate stages of CNN (Khan et al. 2018a).

- In the human visual system, attention is one of the important mechanisms in capturing information from images. The attention mechanism operates in such a way that it not only extracts the essential information from image but also stores its contextual relation with other components of image (Bhunia et al. 2019). In the future, research may be carried out in the direction that preserves the spatial relevance of objects along with their discriminating features at later stages of learning.

- The learning capacity of CNN is generally enhanced by increasing the size of the network, and it can be done in a reasonable time with the help of the current advanced hardware technology such as Nvidia DGX-2 supercomputer. However, the training of deep and high capacity architectures is still a significant overhead on memory usage and computational resources (Lacey et al. 2016; Sze et al. 2017; Justus et al. 2019). Consequently, we still require many improvements in hardware technology that can accelerate research in CNNs. The main concern with CNNs is the run-time applicability. Moreover, the use of CNN is hindered in small hardware, especially in mobile devices, because of its high computational cost. In this regard, different hardware accelerators are needed for reducing both execution time and power consumption (Geng et al. 2019). Some of the very interesting accelerators are already proposed. For example, Application Specific Integrated Circuits, FPGA, and Eyeriss are well known (Moons and Verhelst 2017). Moreover, different operations have been performed to minimize the hardware resources in terms of chip area and energy requirement, by reducing floating-point precision of operands and ternary quantization or minimizing the number of matrix operations. Now it is also time to redirect research towards hardware-oriented approximation models (Geng et al. 2019).

- Deep CNN has a large number of hyper-parameters such as activation function, kernel size, number of neurons per layers, and arrangement of layers, etc. The selection of hyper-parameters and the evaluation time of a deep network, make parameter tuning quite a difficult job. Hyper-parameter tuning is a tedious and intuition driven task, which cannot be defined via explicit formulation. In this regard, Genetic algorithms can also be used to automatically optimize the hyper-parameters by performing searches both in a random fashion as well as by directing search by utilizing previous results (Young et al. 2015; Suganuma et al. 2017; Khan et al. 2019).

- In order to overcome hardware limitations, the concept of pipeline parallelism can be exploited to scale up deep CNN training. Google group has proposed a distributed ML library GPipe (Huang et al. 2018) that offers a  model parallelism option for training. In the future, the concept of pipelining can be used to accelerate the training of large models and to scale the performance without tuning hyper-parameters.

- In the future, it is expected that the potential of cloud-based platforms will be exploited for the development of computationally intensive CNN applications (Akar et al. 2019; Stefanini et al. 2019). Deep and wide CNNs present a critical challenge in implementing and executing them on resource-limited devices. Cloud computing not only allows dealing with a massive amount of data but also leverages the benefit of high computational efficiency at a negligible cost. World-leading companies such as Amazon, Microsoft, Google, and IBM offer the public cloud computing facilities at high scalability, speed, and flexibility to train resource-hungry CNN architectures. Moreover, the cloud environment makes it easy to configure libraries both for researchers and new practitioners.

- CNNs are mostly used for image processing applications, and therefore, the implementation of the state-of-the-art CNN architectures on sequential data requires the conversion of 1D-data into 2D-data. Due to the good feature extraction ability and efficient computations with the limited number of parameters, the trend of using 1D-CNNs is being promoted for sequential data (Vinayakumar et al. 2017; Madrazo et al. 2019).

- Recently high energy physicists at CERN have also been utilizing the learning capability of CNN for the analysis of particle collisions. It is expected that the use of ML and specifically that of deep CNN in high energy physics will grow (Aurisano et al. 2016; Madrazo et al. 2019).

# 8    Conclusion

CNN has made remarkable progress, especially in image processing and vision-related tasks, and has thus revived the interest of researchers in ANNs. In this context, several research works have been carried out to improve CNN's performance on such tasks. The advancements in CNNs can be categorized in different ways, including activation, loss function, optimization, regularization, learning algorithms, and innovations in architecture. This paper reviews advancement in the CNN architectures, especially based on the design patterns of the processing units and thus has proposed the taxonomy for recent CNN architectures. In addition to the categorization of CNNs into different classes, this paper also covers the history of CNNs, its applications, challenges, and future directions.

The learning capacity of CNN is significantly improved over the years by exploiting depth and other structural modifications. It is observed in recent literature that the main boost in CNN performance has been achieved by replacing the conventional layer structure with blocks. Nowadays, one of the paradigms of research in CNN architectures is the development of new and effective block architectures. The role of a block in a network can be that of an auxiliary learner. These auxiliary learners either exploit spatial or feature-map information or even boost input channels to improve performance. These blocks play a significant role in boosting CNN performance by making problem-aware learning.

Moreover, the block-based architecture of CNN encourages learning in a modular fashion and thereby, making architecture simpler and more understandable. The concept of the block being a structural unit is going to persist and further enhance CNN performance. Additionally, the idea of attention and exploitation of channel information, in addition to spatial information, is expected to gain more importance.

## Acknowledgments

The authors would like to thank Pattern Recognition lab at DCIS, and PIEAS for providing them computational facilities. The authors express their gratitude to M. Waleed Khan of PIEAS for the detailed discussion related to the Mathematical description of the different CNN architectures.